\documentclass[preprint,12pt,3p]{elsarticle}

\usepackage{graphicx}
\usepackage{amssymb}
\usepackage{amsthm}
\usepackage{amsmath}
\DeclareMathOperator*{\argmax}{arg\,max}

\usepackage{multirow}
\usepackage{multicol}
\usepackage{float}
\usepackage{subcaption}

\usepackage{xcolor}
\usepackage{lineno}

\journal{ }

\begin{document}

\begin{frontmatter}

\title{Modeling Heterogeneity in Mode-Switching Behavior \\ Under a Mobility-on-Demand Transit System: \\ An Interpretable Machine Learning Approach}



\author[gt-label]{Xilei Zhao}
\ead{xilei.zhao@isye.gatech.edu}
\cortext[cor1]{Corresponding author.}

\author[um-label]{Xiang Yan\corref{cor1}}
\ead{jacobyan@umich.edu}

\author[gt-label]{Pascal Van Hentenryck}
\ead{pvh@isye.gatech.edu}

\address[gt-label]{H. Milton Stewart School of Industrial and Systems Engineering, Georgia Institute of Technology}

\address[um-label]{Taubman College of Architecture and Urban Planning, University of Michigan}

\begin{abstract}
Over the last decade, machine learning has been applied to many
engineering and scientific disciplines, often achieving high
predictive accuracy. Recent years, however, have witnessed an increased
focus on interpretability and the use of machine learning to inform
policy analysis and decision making.
This paper applies machine learning to examine travel behavior and, in particular, on
modeling changes in travel modes when individuals are presented with
a novel (on-demand) mobility option. It addresses the following question: Can machine learning be applied to model
individual taste heterogeneity (preference heterogeneity for travel modes and response heterogeneity to travel attributes) in travel mode choice?

This paper first develops a high-accuracy
classifier to predict mode-switching behavior under a hypothetical Mobility-on-Demand
Transit system (i.e., stated-preference data), which represents the case study underlying this
research. We show that this classifier naturally captures individual heterogeneity available in the data. Moreover, the
paper derives insights on heterogeneous switching behaviors through
the generation of marginal effects and elasticities by current travel
mode, partial dependence plots, individual conditional expectation
plots. The paper also proposes two new model-agnostic interpretation
tools for machine learning, i.e., conditional partial dependence plots and
conditional individual partial dependence plots, specifically designed
to examine response heterogeneity. The results on the case study show that the
machine-learning classifier, together with model-agnostic
interpretation tools, provides valuable insights on travel mode
switching behavior for different individuals and population segments.
For example, the existing drivers are more sensitive to additional pickups
than people using other travel modes, and current transit users are generally willing to share rides but reluctant to take any additional transfers. These insights can be used to inform
transportation planning and the design of new mobility services.
\end{abstract}


\end{frontmatter}



\section{Introduction}
\label{Intro}


The popularity of machine learning is continuously increasing in
transportation, with applications ranging from traffic engineering to
travel demand forecasting and pavement material modeling, to name just
a few. In general, machine learning achieves greater predictive
accuracy compared to traditional methods. However, there is also a
recognition in the field that a single metric, such as the predictive
accuracy or mean squared error, provides an incomplete description of
the complexities arising in the real world
\citep{doshi2017towards}. As a result, in many disciplines, increased
attention is being devoted to the interpretability of machine-learning
results.

This paper applies machine learning to model travel behavior and, in
particular, to model the switching from traveler's usual commute travel modes to a
novel Mobility-on-Demand (MOD) option. Previous studies have applied Machine learning to model travel behavior
\citep[e.g.,][]{xie2003work, tang2015decision,hagenauer2017comparative, zhao2018modeling}. 
These studies often find that machine learning can
predict individual travel-mode choice more accurately than
traditional random utility models, the de facto standard in travel-behavior
modeling. Nonetheless, previous work rarely discuss how to interpret
machine-learning models and derive behavioral insights from the model outputs in order to
inform transportation planning and policymaking. These are areas where random utility models
typically excel due to their microeconomic foundations and the natural interpretations of the utility functions.

This paper extends the applications of machine-learning methods in
travel-mode modeling from a mere focus on prediction to detailed
behavioral interpretations. Its primary goal is to derive behavioral
insights from machine-learning methods that can then be used to inform
transportation planning and policy intervention. More specifically,
the paper examines individual taste
heterogeneity, an essential research topic in travel behavior modeling
that has been a primary focus within the random utility framework
\citep[e.g.,][]{srinivasan2003analyzing,vij2016and,bhat2000incorporating,li2016incorporating,bhat2016allowing}.
To our knowledge, the application of interpretable machine
  learning to model heterogeneity in travel behavior is
  novel and is largely absent from the existing literature.





This paper provides a case study of applying machine learning to model
heterogeneity in mode switching behavior, in the context of
a proposed MOD Transit system that integrates fixed-route services and
on-demand shuttles \citep{TS2017}. It first derives a high-fidelity
predictive model based on machine learning. The paper then shows that
the best-performing machine-learning model (Boosting trees [BOOST]) can naturally segment the entire population and capture heterogeneous mode-switching behavior.
Specifically, the paper demonstrates that behavioral insights can be
revealed through the generation of marginal effects and elasticities for 
different market segments, partial dependence plots
(PDPs), and individual conditional expectation (ICE) plots. Moreover,
the paper proposes two new concepts, conditional partial dependence
plots (CPDPs) and conditional individual partial dependence plots
(CIPDPs), which allows machine learning to capture taste heterogeneity between and across different market segments. As the results of the case study suggest, these novel tools can generate valuable insights regarding the potential adoption of the MOD Transit system and the impact of the underlying design decisions.

The rest of the paper is organized as follows. Section 2 summarizes
the research background and includes three major parts, i.e., machine-learning applications in travel behavior modeling, modeling heterogeneity in travel behavior, and interpretable
machine learning.  Section 3 describes the methodological framework
underlying this paper: It discusses the use of machine learning for
travel choice modeling and available tools for interpreting the
heterogeneous mode switching behavior. In addition, it presents two
novel tools, CPDPs and CIPDPs, which allow for the generation of more
behavioral insights by applying market segmentation. Section 4
describes a case study, which includes the data description, the
specification of machine-learning classifiers, and the training,
validation, and testing procedure. Section 5 presents results on the
predictive capability of machine learning, PDPs and CPDPs, ICE plots
and CIPDPs, and marginal effects and elasticities for different market segments. Finally, Section 6 summarizes the findings,
identifies the benefits and limitations of the proposed approach, and
suggests future research directions.

\section{Research Background}

\subsection{Machine Learning Applications in Travel Behavior Modeling}

In recent years, researchers started to apply machine-learning methods
to model individual travel behavior \citep[e.g.,][]{xie2003work,tang2015decision, hagenauer2017comparative,wang2018machine}. For example, \citet{xie2003work} applied decision
trees and neural networks (NN) to model individual mode choice and showed
that these two models could improve the predictive accuracy of the
multinomial logit model.  Later, \citet{tang2015decision} also applied
decision trees to model travel mode switching behavior and obtained
higher accuracy than logit models in most cases.  More recently,
\citet{hagenauer2017comparative} provided a comparison of various
machine-learning classifiers for modeling travel mode choice, focusing
primarily on predictive accuracy. \citet{wang2018machine} found that the extreme gradient boosting model substantially exceeded the prediction performance of the multinomial logit model when modeling travel mode choices.

Even though machine learning has demonstrated its strength in mode-choice
predictions, there has been little discussion regarding interpreting the machine learning
model outputs to extract behavioral insights. That is to say, knowledge on
the underlying decision rules that machine learning uses for travel-behavior prediction is lacking.
\citet{zhao2018modeling} provides an early probe into this issue. They conducted a comprehensive comparison between machine learning and random utility models and found that that machine learning
can not only offer a more flexible modeling structure and achieve higher predictive
accuracy than traditional logit models, but also produce comparable behavioral outputs such as marginal effects and elasticities.

\subsection{Modeling taste Heterogeneity in Travel Behavior}

Travel behavior heterogeneity has been extensively studied within
the random utility modeling framework. There are two types of taste heterogeneity: one is preference heterogeneity, which refers to the varying levels of preference for different travel modes across individuals; the other is response heterogeneity, which indicates travelers' varying levels of sensitivity to changes in travel attributes. From a modeler's perspective, taste
heterogeneity can be divided into two parts, i.e., observed and
unobserved. \textit{Observed heterogeneity} can be captured by
introducing observed individual socio-demographic or behavioral
characteristics as alternative specific variables and/or by capturing
interactions between level-of-service variables and observed
individual features, e.g., by applying a market segmentation approach
or adding interaction terms \citep{bhat2000incorporating}. The market segmentation approach means to select subgroups from a population in advance based on
the known characteristics and declare them as ``segments,'' aiming at
analyzing a manageable number of groups that share well-defined
underlying features and generating more creative and better-targeted
policies for different groups \citep{anable2005complacent}. On the
other hand, \textit{unobserved heterogeneity} is usually caused by
unobserved individual features, such as individuals' intrinsic bias towards different travel modes or their varying degrees of sensitivity to level-of-service attributes \citep{bhat2000incorporating}. To
account for unobserved heterogeneity, researchers usually applied
mixed logit models to fit random coefficients for explanation
variables and/or to fit flexible
error terms (the independent Gumbel distributed part plus the
correlated normal distributed part) to account for the correlations
over alternatives \citep{li2016incorporating}.

With strong theoretical foundations, random utility models have been
widely applied to model heterogeneity
\citep[e.g.,][]{srinivasan2003analyzing,vij2016and,bhat2000incorporating,li2016incorporating,bhat2016allowing}. For
instance, \citet{bhat2000incorporating} took into account observed and
unobserved heterogeneity for modeling urban work travel mode choice.
\citet{srinivasan2003analyzing} proposed a dynamic kernel logit
formulation to analyze heterogeneity and unobserved structural effects
in route-switching behavior. To capture the disaggregate
decision-making more accurately, researchers have used the integrated
choice and latent variable models in assessing heterogeneous travel
behavior \citep{vij2016and}.

Nevertheless, since random utility models often have low prediction
performance \citep{hagenauer2017comparative,zhao2018modeling}, one may
think that the behavioral insights generated from them are less
trustworthy than those based on models with higher predictive
accuracy. Moreover, recent studies have suggested that, unlike
conventional random utility models that require extensive modeling effort and
domain knowledge to accommodate individual heterogeneity, machine learning models
can account for it automatically. For example,
\citet{lheritier2018airline} compared the random forest (RF) model
with the latent class multinomial logit model through a series of
experiments, and found that the RF model has the ability of segmenting population
groups (with heterogeneous tastes) automatically. Therefore, it would be
worthwhile to study the application of interpretable machine learning
techniques to examine individual taste heterogeneity.

\subsection{Interpretable Machine Learning}

Interpretable machine learning is receiving increasing attention in
recent years \citep[e.g.][]{murdoch2019interpretable,
  du2018techniques, doshi2017towards,
  molnar,zhao2017causal,wager2018,athey2017beyond}.
In particular, \citet{murdoch2019interpretable} defined interpretable machine
learning as applying machine-learning methods to extract
\textit{relevant} knowledge about the domain relationships contained in the data. 

The methods for machine-learning interpretability may be
divided into two categories, i.e., intrinsic and post hoc. As
discussed in \citet{molnar}, \textit{intrinsic} interpretability
usually refers to relatively simple machine-learning models that are
considered interpretable due to their simple model structure, such as
linear regression, logit models, and decision trees. On the other
hand, \textit{post hoc} interpretability of a machine-learning model
is achieved by applying interpretation methods after its training and
application. The machine-learning interpretation methods can also be
divided into model-specific or model-agnostic. \textit{Model-specific}
methods can only be applied to a specific class of models, while
\textit{model-agnostic} methods can be applied to any machine-learning
models after training. Since model-agnostic interpretation methods
have lots of flexibility compared to their dedicated counterparts and
can provide consistent interpretability criteria for any model class,
this paper focuses on applying and inventing model-agnostic interpretation methods
to explain individual travel behavior.

One of the most prevalent model-agnostic methods is the PDP, 
which presents the dependence between the
response variable and a set of input features, marginalizing over the
values of the remaining features \citep{friedman2001greedy}. As an
extension of PDP, \citet{goldstein2015peeking} proposed another
model-agnostic method---the ICE
plots---to reveal the potential individual heterogeneity by generating
one curve per observation that presents how its prediction evolves
when a feature changes. Notably, PDPs and ICE plots were recently proved
to be effective to reveal causal inference between input features and
the response variable \citep{zhao2017causal}. In addition, \citet{wager2018}
assessed estimation and inference of heterogeneous treatment effects
using RF. However, PDPs and ICE curves focus on all the
instances under evaluation without looking at the specific groups of
people to gain additional insights on taste heterogeneity across different population groups. Therefore, this paper introduces two new model-agnostic methods as an extension of PDP
and ICE plots, in order to better uncover the taste heterogeneity within and across different population segments.

Some model-agnostic methods have been applied in some of the existing
literature of travel behavior modeling, such as
\citet{hagenauer2017comparative,wang2018machine}. However,
their discussion mainly focused on variable importance (measures the
importance of a feature for prediction). Our prior work
\citep{zhao2018modeling} showcased how to apply PDPs to evaluate the
marginal impacts of the selected features. However, to our
knowledge, no existing work focuses on applying existing
machine-learning interpretation tools or developing new ones to reveal
heterogeneous travel behavior for informing the design of MOD Transit and policy analysis.





\section{Methodological Framework}
\label{Mtd}

This paper is interested in the following research question: What
factors and how they shape individual willingness to switch to a new transportation
mode? The paper approaches this question using the following
methodological framework. It assumes the availability of a
longitudinal dataset that captures individual travel mode choice
before and after a new mobility option is introduced. If the new
mobility service is not yet deployed, a stated-preference (SP) survey
can be conducted to capture individual preferences for the new travel
mode.  The data can be represented as $\{\boldsymbol{x}_i, y_i\}_{i =
  1}^N$, where $\boldsymbol{x}_i = [x_{i1}, ..., x_{ip}]$ is a vector
of $p$ features for individual $i$ and $y_i$ is the response
variable. The features $\boldsymbol{x}_i$ often include socio-economic
and demographic information, travel preference, current travel mode,
and the level-of-service variables for each travel mode under
evaluation. The response variable $y_i$ is binary: Value 0 indicates
that individual $i$ stays with her current mode and value 1 indicates
that individual $i$ switches to the new travel mode.

\subsection{Choice Modeling with Machine Learning}

Choice modeling with machine learning is typically approached as a
{\em classification} problem
\citep{xie2003work,hagenauer2017comparative,zhao2018modeling}. Commonly used classification models
include logistic regression (logit)
\citep{friedman2001elements}, Support Vector Machines (SVM)
\citep{cortes1995support}, RF
\citep{breiman2001random}, and BOOST
\citep{friedman2001greedy}. The classification approaches can be divided into two categories, i.e., soft and hard classification
\citep{wahba2002soft, liu2011hard}. \textit{Soft classification}
estimates the class conditional probabilities first and then predicts
the class based on the largest estimated probability
\citep{liu2011hard}. On the other hand, \textit{hard classification}
predicts the class labels directly without estimating intermediate class
probabilities. Most prior work in machine learning for choice modeling
are based on hard classification \citep[e.g.][]{xie2003work,
  tang2015decision, wang2018machine}. However, it is
more natural to use class probabilities (estimated by 
soft classification) to predict market shares,
especially when the sample size is small. For example, if the sample
population contains a hundred people for which the predicted switching
probabilities are all 0.51, a hard classification will predict
switching in 100\% of the cases, while a soft classification would
concludes that switching occurs in 51\% of the cases.

This paper treats the prediction of travel mode switching as a soft
classification problem. The classifier
$f(\boldsymbol{x}|\hat{\boldsymbol{\theta}})$ (where
$\hat{\boldsymbol{\theta}}$ is an estimated parameter or
hyperparameter vector) maps the features $\boldsymbol{x}$ to
the response variable $\boldsymbol{y}$ using the predictions of class
probabilities, i.e.,
\begin{equation}
    \hat{y}_i = \argmax_{k \in \{0, 1\}}  \hat{p}_{i,k}
\end{equation}
where $\hat{p}_{i,k}$ is the predicted choice probability for class
$k$ ($k \in \{0, 1\}$) of observation $i$ ($i = 1, 2, ..., N$).

The predictive accuracy at the individual level of different classifiers is evaluated using the formula
\begin{equation}
\sum_{i=1}^N I_{y_i}(\hat{y}_i)/N,
\end{equation}
where $I_{y_i}(\hat{y}_i)$ is an indicator function which equals to 1 when $\hat{y}_i = y_i$. 

The predicted market share for class $k$ is computed using the formula
\begin{equation}
   Q_k(\boldsymbol{x}|\hat{\boldsymbol{\theta}}) = \frac{1}{N}\sum_{i=1}^N \hat{p}_{ik}
\end{equation}
that averages the predicted choice probabilities for all the
instances. The predictive accuracy at the aggregate level is evaluated
using the L1-Norm, i.e.,
\begin{equation}
    \sum_{k=0}^1 \Big| Q_k^* - Q_k(\boldsymbol{x}|\hat{\boldsymbol{\theta}}) \Big|,
\end{equation}
where $Q_k^*$ represents the observed market share for choice $k \in \{0, 1\}$.

The training, validation, and testing of the classifiers, as well as
the model selection, are discussed in detail in Subsection 4.3.

\subsection{Interpreting Heterogeneous Mode Switching Behavior}

Soft classification helps the interpretation of heterogeneous
switching behavior. This section reviews some existing interpretation
techniques, including PDPs and their extensions in ICE curves. In
addition, it proposes two new tools for interpretation, CPDPs and
CIPDPs.  Finally, this section reports marginal effects and
elasticities for market segments in order to further understand
individual taste heterogeneity.

\subsubsection{Partial Dependence Plots (PDPs) and Individual Conditional Expectation (ICE) Plots}

PDPs, first proposed by \citet{friedman2001greedy}, are one of the
most popular model-agnostic interpretation tools for machine-learning
models. Assume that we are interested in determining the effect of a
set $S \subset \{1, ..., p\}$ of features on the prediction outcomes
and let $C$ be the complement of $S$ (i.e., $C = \{1, ..., p\}
\setminus S$). The \emph{partial dependence} of classifier $f$ on
$\boldsymbol{x}_{S}$ is defined as
\begin{equation}
    f_{S}(\boldsymbol{x}_{S}) = \mathbb{E}_{\boldsymbol{x}_{C}}f(\boldsymbol{x}_{S}, \boldsymbol{x}_{C}|\boldsymbol{\theta}).
\end{equation}
In practice, Eqn. (5) is estimated by computing
\begin{equation}
  \hat{f}_{S}(\boldsymbol{x}_{S}) = \frac{1}{N}\sum_{i = 1}^N f(\boldsymbol{x}_{S}, \boldsymbol{x}_{C_i}|\hat{\boldsymbol{\theta}}),
\end{equation}
where $\boldsymbol{x}_{C_i} (i = 1, ..., N)$ represent the values of
$\boldsymbol{x}_{C}$ for each instance in the training set. The PDP
evaluates the influence of $\boldsymbol{x}_{S}$ on $\hat{f}$ after
marginalizing over all the other features. For soft classifiers, the
PDP displays the class probability (e.g., the switching probability)
for each possible value of $\boldsymbol{x}_{S}$. As discussed in
\citet{friedman2001greedy}, the PDP can be a useful summary of the
impact of the chosen subset of features when the interactions between the chosen features and the
remaining features are weak. However, when the interactions are
strong, the PDPs may obscure a heterogeneous relationship created by the interactions \citep{goldstein2015peeking}.



To complement PDPs, \citet{goldstein2015peeking} proposed ICE plots to
capture the potential individual heterogeneity. Instead of plotting the average partial dependence on
the predicted response, the ICE curve generates an estimated
conditional expectation curve
\begin{equation}
\hat{f}_{S}^i(\boldsymbol{x}_{S}) = f(\boldsymbol{x}_{S}, \boldsymbol{x}_{C_i}|\hat{\boldsymbol{\theta}}),
\end{equation}
for each instance $C_i$ $(i=1,\ldots,N)$ in the dataset. The average
of all the ICE plots is the corresponding PDP for the selected feature
\citep{molnar}. As ICE plots generate individual-specific curves, it
can be used directly to understand taste
heterogeneity. 

PDPs and ICE plots are easy to implement and provide clear
interpretations of a classifier. In particular, ICE plots are capable
of revealing individual heterogeneity (by producing
individual-specific curves), which is an important topic in travel
behavior modeling. Furthermore, as discussed by
\citet{zhao2017causal}, PDPs and ICE plots may reveal causal
relationships if $f$ is an accurate classifier and domain knowledge
supports the underlying causal structure.


It is important to point out that PDPs can be thought as a counterpart
in machine learning relative to the beta coefficients in a logit model. A PDP graphically illustrates the relationship between an input feature and the response variable. For linear machine-learning models, the PDP would be a straight line, whose slope is equivalent to a beta coefficient; for nonlinear models, the PDP would be a curvy line with different beta coefficients (i.e. tangent of the line) at different data points. To compare the results of PDP from nonlinear machine-learning models with the beta coefficient in a logit model, we propose to use a straight line to approximate the relationship displayed in the PDP by calculating a single slope for the PDP using:
\begin{equation}
    \frac{\hat{p}_1^{\text{max}} - \hat{p}_1^{\text{min}}}{x_{\text{max}} - x_{\text{min}}}
\end{equation}
where $x_{\text{max}}$ and $x_{\text{min}}$ are the maximum and
minimum values of a selected feature, and $\hat{p}_1^{\text{max}}$ and
$\hat{p}_1^{\text{min}}$ are the corresponding estimated switching
probabilities at $x_{\text{max}}$ and $x_{\text{min}}$ respectively.
Note that the beta coefficients of a logit model are in the logit
region (or log-odds region), while the the slopes of PDPs are in the
probability domain.

\subsubsection{Conditional Partial Dependence Plots (CPDPs) and Conditional Individual Partial Dependence Plots (CIPDPs)}

This paper also proposes two new model-agnostic tools for improving
the interpretations of classifiers and revealing additional insights
in presence of individual heterogeneity: CPDPs and CIPDPs.  The key
idea behind CPDPs and CIPDPs is to group instances into subpopulations
based on some observed features (i.e., a market segmentation
approach). These plots further allow researchers to examine response heterogeneity across different population groups. 

This paper uses the observed socio-demographic or behavioral features
to segment the market in order to examine \textit{observed
heterogeneity}.  For example, the instances can be grouped by
income, gender, or current commuting mode. By plotting the ICE curves
within each market segment and computing the corresponding PDP for the
subpopulation, it is possible to contrast the behaviors of different
population groups and different individuals within these groups.

More formally, consider a categorical feature $g$ and its set $V$ of
possible values. A CPDP conditioned on feature $g$ is defined as
\begin{equation}
f_{S}(\boldsymbol{x}_{S})_{\mid x_g = v} =  \mathbb{E}_{\boldsymbol{x}_{C}}f(\boldsymbol{x}_{S}, \boldsymbol{x}_{C}|\boldsymbol{\theta},x_g = v)
\end{equation}
for each value $v \in V$. This definition can be generalized to a set $\boldsymbol{G}$ of features 
\begin{equation}
f_{S}(\boldsymbol{x}_{S})_{\mid \boldsymbol{x}_{\boldsymbol{g}} = \boldsymbol{v}} =  \mathbb{E}_{\boldsymbol{x}_{C}}f(\boldsymbol{x}_{S}, \boldsymbol{x}_{C}|\boldsymbol{\theta},\boldsymbol{x}_{\boldsymbol{g}} = \boldsymbol{v}).
\end{equation}

\noindent
CIPDPs are the conditional counterpart to the ICE curves. Given a feature $g$
and a value $v$ of interest, CIPDPs will display the ICE curves,
$f_{S}^j(\boldsymbol{x}_{S})$, for those instances $j$ satisfying
$x_{g,j} = v$, $j \in \{1, ..., N\}$. These definitions can be naturally extended to
non-categorical attributes by partitioning their domains.

Compared to PDPs and ICE plots, CIPDPs and CPDPs mainly assess the pre-determined mutually-exclusive subgroups
of the training set. By segmenting the market based on the prior knowledge of the travel behavior theory, CIPDPs and CPDPs are expected to be useful to capture diverse behaviors across and within different population groups. 





\subsubsection{Marginal Effects and Elasticities for Market Segments}

Marginal effects and elasticities are widely-used econometric concepts to indicate the sensitivity of an outcome variable with respect to changes in independent variables. In mode-choice applications, they are defined as the changes in
the choice probability of an alternative in response to a one unit
(percent) change in an independent variable. These tools can be
applied to machine-learning classifiers too, as shown in
\citet{zhao2018modeling}. The marginal effects and arc elasticities of
feature $\boldsymbol{x}_{t}, t = 1, ..., p$, for class $k$ can be
computed as:
\begin{equation}
    M_k(\boldsymbol{x}_t) = \frac{Q_k(\boldsymbol{x}_{-t}, \boldsymbol{x}_t + \delta|\hat{\boldsymbol{\theta}}) - Q_k(\boldsymbol{x}|\hat{\boldsymbol{\theta}})}{|\delta|},
\end{equation}
\begin{equation}
    E_k(\boldsymbol{x}_t) = \frac{[Q_k(\boldsymbol{x}_{-t}, \boldsymbol{x}_t \cdot (1+\delta)|\hat{\boldsymbol{\theta}}) - Q_k(\boldsymbol{x}|\hat{\boldsymbol{\theta}})]/Q_k(\boldsymbol{x}|\hat{\boldsymbol{\theta}})}{|\delta|},
\end{equation}
where $\boldsymbol{x}_{-t} \in \mathbb{R}^{p-1}$ represents the complement set of $\boldsymbol{x}_{t}$ and $\delta$ is a constant. 

In order to uncover heterogeneity across different population groups, one can also compute marginal effects and elasticities for different market segments instead of for the entire population. For instance,
for a categorical feature $g$ and its potential value $v$, the
marginal effects and elasticities can be defined as
\begin{equation}
    M_k(\boldsymbol{x}_{t})_{|x_g = v} = \frac{Q_k(\boldsymbol{x}_{-{t}}, \boldsymbol{x}_{t} + \delta|\hat{\boldsymbol{\theta}},x_g=v) - Q_k(\boldsymbol{x}|\hat{\boldsymbol{\theta}},x_g = v)}{|\delta|},
\end{equation}
\begin{equation}
    E_k(\boldsymbol{x}_{t})_{|x_g=v} = \frac{[Q_k(\boldsymbol{x}_{-{t}}, \boldsymbol{x}_{t} \cdot (1+\delta)|\hat{\boldsymbol{\theta}},x_g=v) - Q_k(\boldsymbol{x}|\hat{\boldsymbol{\theta}},x_g=v)]/Q_k(\boldsymbol{x}|\hat{\boldsymbol{\theta}},x_g=v)}{|\delta|}.
\end{equation}

\noindent
As discussed by \citet{zhao2018modeling}, some machine-learning
classifiers, such as tree-based models, split input feature values at
different nodes. Therefore, the prediction decisions for these models
become insensitive to changes in feature value outside its range. As a
result, this paper only predicts the marginal effects and elasticities
for instances within the data range. Subsection 5.4 provides more details on this topic.

\section{Case Study}


\subsection{The Data}

The case study uses data collected from a SP survey by faculty, staff,
and students at the University of Michigan, Ann Arbor. The survey
first asked the participants to estimate the travel time, cost, and
wait time for their commuting trip from one of the following modes:
\textit{Car}, \textit{Walk}, \textit{Bike}, and \textit{Bus}. Then, by
showing them a new \textit{MOD Transit} system that would replace the
existing transit system, the survey asked them what travel mode (i.e.,
\textit{Car}, \textit{Walk}, \textit{Bike}, and \textit{MOD Transit})
they might choose in a different state-choice experiments (with
different levels of service for MOD Transit). The detailed survey
description with graphical illustrations can be found in
\citet{YAN2018}.

This paper aims at evaluating the factors underlying individuals'
intention to switch to the new MOD Transit system. That is to say,
some travel behavior changes, such as switching from Car to Bike, is not of
interest in this study. Hence, the outcomes
of the state-choice experiments are expressed as binary answers, where
value 0 represents the decision of {\em not switching to MOD Transit}
and 1 denotes the decision of {\em switching to MOD Transit}. {\em MOD Transit} 
is considered as a new travel mode here, that is to say, individuals who currently
use the bus are considered as {\em switch to} MOD Transit if he or she select the new mode.


\begin{table}[!t]
\footnotesize
\caption{Statistics for Features and Response Variable.}
\setlength{\tabcolsep}{7pt}
\centering
\resizebox{1\textwidth}{!}{
     \begin{tabular}{ l | l | l l l l l l }
       \hline
       \textbf{Variable} & \textbf{Description} & \textbf{Category} & \textbf{\%} & \textbf{Min} & \textbf{Max} & \textbf{Mean} & \textbf{SD}  \\[0.5ex]  \hline
       \textit{Response Variable} &&&&&&& \\[0.5ex]
       Switching Choice  & & MOD Transit  & 35.28 &        &       &  & \\
       &   & Not MOD Transit & 64.72  &        &       &  & \\
       & &&&&&& \\
       \textit{Features} &&&&&&& \\[0.5ex]
       TT\_Drive & Travel time of driving (min) &  &  & 2.00 & 40.00 & 15.21 & 6.62
       \\
       TT\_Walk & Travel time of walking (min)  &  &  & 3.00 & 120.00 & 32.30 & 23.08
       \\
       TT\_Bike & Travel time of biking (min) &  &  & 1.00 & 55.00 & 15.34 & 10.45
       \\
       TT\_MOD & Travel time of using MOD transit (min)    &  &  & 6.20 & 34.00 & 18.68 & 4.75
       \\
       Wait\_Time & Wait time for MOD (min) &  &  & 3.00 & 8.00  & 5.00 & 2.07
       \\
       Transfer & Number of transfers in MOD &  &  & 0.00 & 2.00  & 0.33 & 0.65
       \\
       Rideshare & Number of additional pickups in MOD &  &  & 0.00 & 2.00  & 1.11 & 0.82
       \\
       Income& Income level
       &  &  & 1.00 & 6.00  & 1.93 & 1.34 \\
       Bike\_Walkability& Importance of bike- and walk-ability   &  &  & 1.00 & 4.00  & 3.22 & 0.95 \\
       MOD\_Access& ease of access to MOD  &  &  & 1.00 & 4.00  & 3.09 & 1.02  \\
       CarPerCap & Car per capita  &  &  & 0.00 & 3.00  & 0.53 & 0.48
       \\
       Female & Female or Male  & Female   & 56.32   & & & & \\
       & & Male & 43.68  & & & & \\
       Student & Students or faculty/staff  & Student & 73.52 & & & & \\
       && Faculty or staff & 26.48 & & & & \\
       Current\_Mode\_Car  & Current travel mode is Car or not & Car & 16.68 & & & & \\
       && Not Car & 83.32 & & & & \\
       Current\_Mode\_Walk  & Current travel mode is Walk or not & Walk & 40.41 & & & & \\
       && Not Walk & 59.59 & & & & \\
       Current\_Mode\_Bike  & Current travel mode is Bike or not & Bike & 8.25 & & & & \\
       && Not Bike & 91.75 & & & & \\
       [1ex]
       \hline
    \end{tabular}
    }
     \label{tab:var}
\end{table}

There were 8,141 data points collected from 1,163 individuals. The
statistics for 16 features and the response variable are given in
Table \ref{tab:var}. The feature ``Current\_Mode\_Bus'' 
(indicates whether the current travel mode is Bus) is not included for
analysis, since this feature is completely correlated with the other
three features: ``Current\_Mode\_Car,'' ``Current\_Mode\_Walk,'' and
``Current\_Mode\_Bike,'' with the sum of the four features equal to 1. All the variables
have the variance inflation factor less than 5 (a threshold commonly referenced), which suggests that 
multicollinearity is not a concern.

\begin{figure}[!t]
    \centering
    \includegraphics[width=9cm]{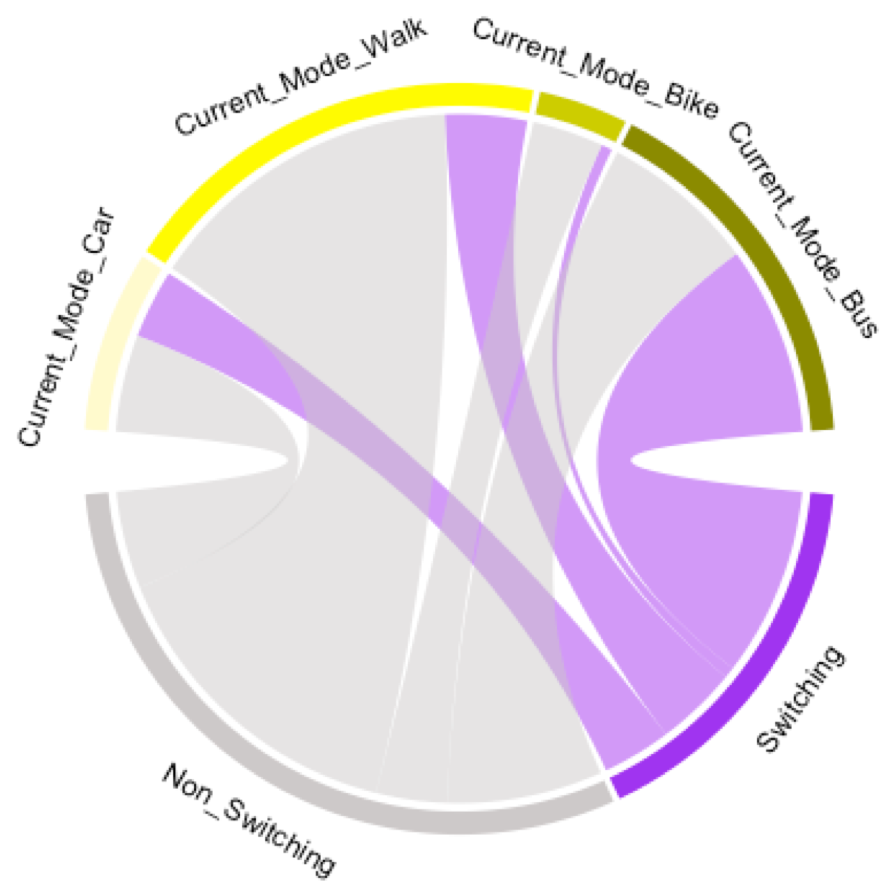}
    \caption{Chord Diagram of Switching Choice Versus Current Travel Mode Choice.}
    \label{fig:Chord}
\end{figure}


Based on the observed socio-demographic and behavioral
characteristics, there are different ways to segment the market to
study individual heterogeneity. Some possible
market segmentation approaches include grouping people by different
income levels, car ownership, gender, or commonly-used commuting
modes. In particular, our previous study \citep{zhao2018modeling} has
found out that, for the multinomial logit model, the dummy variables
of the current travel modes are positive for all the four modes and
statistically significant for Car, Walk, and Bike. These results
indicate that travelers may present an inertia to their current
modes. Moreover, as an exploratory analysis, Figure \ref{fig:Chord}
presents a chord diagram illustrating graphically the
inter-relationships between switching decisions and current travel
modes. It is obvious that different market segments show different
ratios of switching. Therefore, as an illustration, this paper chooses
to segment the market using the current travel mode and shows how to
apply interpretable machine learning to reveal individual heterogeneity in mode-switching behavior under a proposed MOD Transit system. 





\subsection{Machine-Learning Classifiers}

Seven soft machine-learning classifiers are selected for comparison:
They include logit, naive Bayes (NB), classification and regression trees (CART), bagging trees (BAG),
BOOST, RF, and NN. This subsection briefly summarizes these classifiers: It may be
skipped by those familiar with these techniques.

The logit model is arguably the most popular classifier to predict a
binary outcome (i.e., switching or not switching)
\citep{friedman2001elements}. The logit model assumes a linear
relationship between the log-odds and input features, facilitating the
interpretation of the results and the derivation of policy
interventions. However, the linear nature of this classifier may not
reflect the true relationship between the log-odds and input features,
and the logit model may exhibit a relatively low predictive accuracy.

The NB model is also widely-used for classification. It assumes that
all features are independent \citep{mccallum1998comparison}. NB is
easy to construct and is often used as a baseline classifier. However,
its assumption is often violated in practice, making it sensitive to
highly correlated features.

The CART model builds a classification tree, where each internal node
partitions the data based on the value of a selected feature and
leaves capture a class decision (e.g., switching or not switching in
the case study) \citep{breiman2017classification}. The decision tree
is susceptible to overfitting \citep{quinlan2014c4} and pruning
techniques can control its complexity. In this study, the trees are
grown without pruning, since they are very simple, i.e., less than 5-6
leaves in most cases.

\citet{breiman1996bagging} and \citet{friedman2001elements} proposed a
number of tree-based ensemble methods to overcome the limitations of
CART classifiers by providing more accurate, stable, and robust
models. There are three major ensemble models, including BOOST, BAG,
and RF. Specifically, for classification problems, BOOST models create
a sequence of decision trees and each successive tree is designed to
improve the predictive accuracy of its predecessor.  The final
prediction of the BOOST model is the weighted voting among all
trees. Compared to CART, the BOOST model usually produces a higher
predictive accuracy. This study applies gradient boosting
\citep{friedman2001greedy}. 500 trees are used, with shrinkage
parameter set to 0.062 and interaction depth to 45. The minimum number
of observations in leaves is 10. 
On the other hand, BAG and RF train a set of trees using
bootstrapping (i.e., sampling with replacement)
\citep{breiman1996bagging,ho1998random}. The only difference between
BAG and RF is that BAG uses all features to train the trees while RF
only selects a random subset of all the features to train the
trees. For a classification problem, multiples decision trees are
trained, and the majority voting among all the trees is the prediction
outcome. By using bootstrapping, the BAG model can reduce the variance
and overfitting problems of a single decision tree. However, by
assuming variable independence, BAG cannot reduce the variance for
correlated features. By contrast, RF may overcome this limitation and
reduce the variance between correlated trees. For the BAG model, 500
classification trees are trained, with each tree grown without
pruning. For the RF model, 600 trees are used and 14 randomly selected
variables are considered for each split at internal nodes.



A basic NN model with three layers of nodes (each node is binary) was
also considered. NN has an input layer, a hidden layer, and an output
layer. Each node connection between adjacent layers has a weight. The
hidden layer of the NN model can help measure nonlinear relationships
between features and the response variable. However, NN is also
susceptible to overfitting. In the paper, a NN model with a single
hidden layer of 14 units is used. The connection weights are trained
by back propagation with a weight decay constant of 0.1.


All the modeling and analyses are conducted in R, using following
packages: \textit{stats} \citep{R}, \textit{e1071} \citep{e1071},
\textit{tree} \citep{tree}, \textit{gbm} \citep{gbm},
\textit{randomForest} \citep{RF}, and \textit{nnet} \citep{stats}.

\subsection{Training, Validation, and Testing for the Classifiers}

\begin{figure}[!t]
    \centering
    \includegraphics[width=13cm]{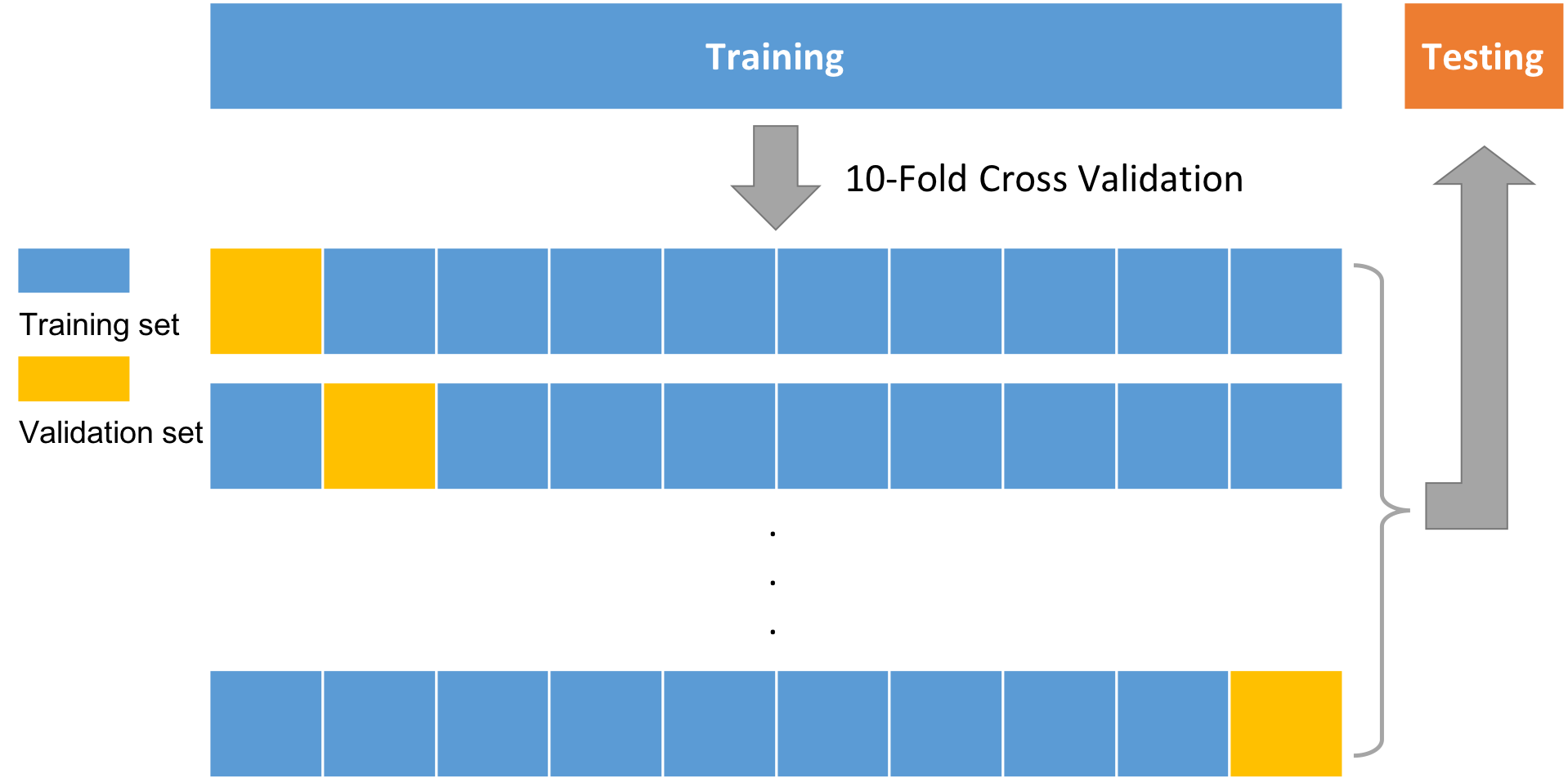}
    \caption{The Training, Validation, and Testing of the Classifiers.}
    \label{fig:CV}
\end{figure}

To identify the best-performing model, this study trained, validated,
and tested the seven machine learning classifiers using the
procedure illustrated in Figure \ref{fig:CV}. First, 10\% of the data
is randomly drawn for the testing set using stratified sampling based
on the current mode choices. In other words, the entire data set is
divided into four mutually-exclusive subpopulations based on their
current mode choices, i.e., Car, Walk, Bike, and Bus. Within each
subpopulation, 10\% samples are randomly drawn for testing.

Next, the remaining 90\% of the data is used to train the model and
tune hyperparameters using a 10-fold cross validation. To conduct the
10-fold cross validation, the training set is first randomly split
into 10 disjoint subsets. Then, one subset is held out for validation
while the remaining nine subsets are used for training the machine
learning classifiers. After repeating the same process for each one of
the 10 subsets, the validation outcomes for each classifier are
average to obtain a mean estimate of the performance metric (e.g., the
predictive accuracy). The model with the best performance metric is
selected and is fitted on the entire training set (i.e., 90\% of the
entire data set). Finally, the selected model is applied to the
testing set to provide an unbiased evaluation of its predictive
capability.

\section{Model Interpretation}
\label{MC}

\subsection{Predictive Accuracy}

\begin{figure}[!t]
    \centering
    \includegraphics[width=12cm]{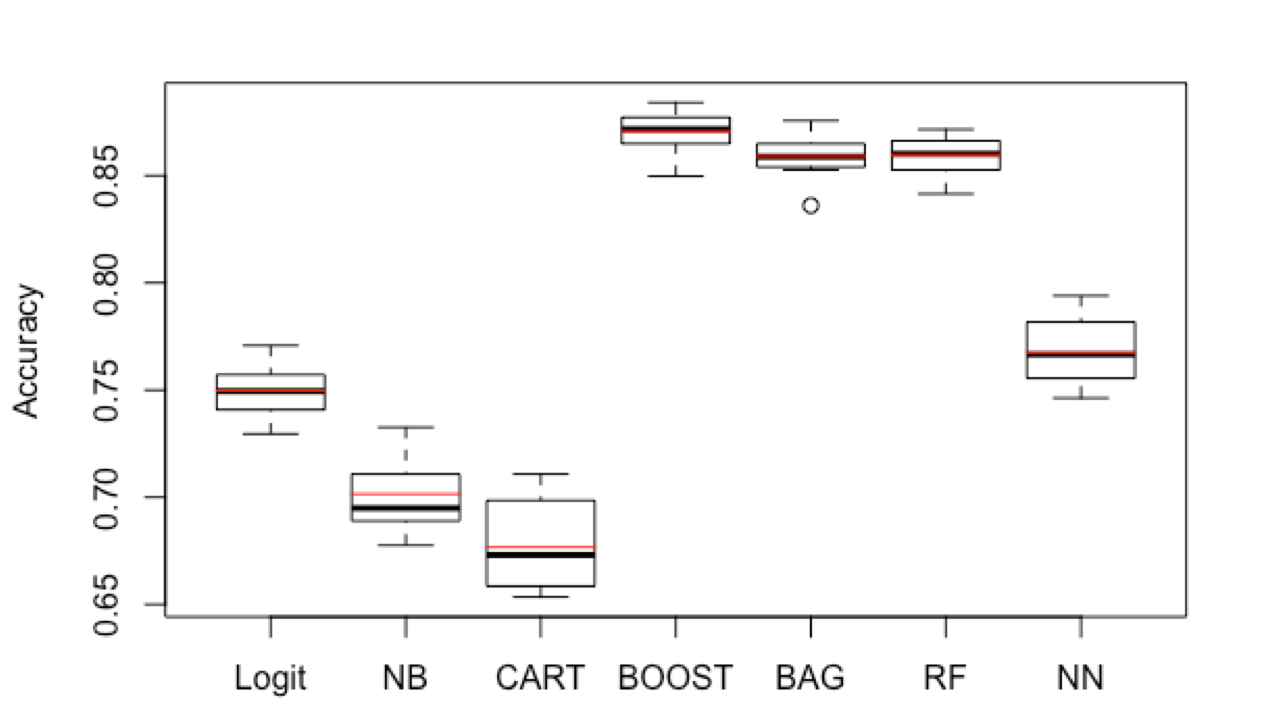}
    \caption{Accuracy of the Classifiers: Boxplots of Cross-Validation
      Results (Red Lines are Mean Values).}
    \label{fig:BoxCV}
\end{figure}

The cross-validation results of seven machine learning classifiers are
shown in Figure \ref{fig:BoxCV}. With respect to the mean predictive
accuracy (red lines in Figure \ref{fig:BoxCV}), BOOST has the best
outcome (0.871), followed by RF (0.860) and BAG (0.859). By contrast,
the mean accuracy of the logit model is only 0.750. The
model with the worst performance is CART (0.677).

\begin{table}[!t]
\centering
\begin{tabular}{c|c|c|c}
\hline
\multicolumn{2}{c|}{\textbf{Mode/Outcome}}            & \textbf{Accuracy} & \textbf{L1-Norm}                  \\ \hline
\multicolumn{2}{c|}{Overall}         & 0.893   & \multirow{3}{*}{0.02695} \\ \cline{1-3}
\multicolumn{2}{c|}{Switching}       & 0.808   &                          \\ \cline{1-3}
\multicolumn{2}{c|}{Non-switching}   & 0.939   &                          \\ \hline
\multirow{3}{*}{Car}  & Overall       & 0.869   & \multirow{3}{*}{0.00213} \\ \cline{2-3}
                      & Switching     & 0.836   &                          \\ \cline{2-3}
                      & Non-switching & 0.890   &                          \\ \hline
\multirow{3}{*}{Walk} & Overall       & 0.904   & \multirow{3}{*}{0.04375} \\ \cline{2-3}
                      & Switching     & 0.620   &                          \\ \cline{2-3}
                      & Non-switching & 0.981   &                          \\ \hline
\multirow{3}{*}{Bike} & Overall       & 0.946   & \multirow{3}{*}{0.00433} \\ \cline{2-3}
                      & Switching     & 0.714   &                          \\ \cline{2-3}
                      & Non-switching & 0.979   &                          \\ \hline
\multirow{3}{*}{Bus}  & Overall       & 0.882   & \multirow{3}{*}{0.02525} \\ \cline{2-3}
                      & Switching     & 0.890   &                          \\ \cline{2-3}
                      & Non-switching & 0.873   &                          \\ \hline
\end{tabular}
\caption{Predicted Accuracy for all Intances and for Instances
  Segmented by Current Mode Choice in the \textit{Testing} Set.}

\label{tab:accuracy}
\end{table}

The BOOST model is thus selected and fitted on the entire training
set. Then, the BOOST model is evaluated on the testing set and the
results are presented in Table \ref{tab:accuracy}. The results show
that the overall accuracy at the individual level is 0.893 with an
accuracy of 0.808 for switching and 0.934 for non-switching
predictions. By using soft classification to predict market shares, 
the overall L1-Norm is 0.02695, indicating that the least
absolute deviation of the market share prediction is less than 3\%.

Table \ref{tab:accuracy} also reports the predictive capabilities of
the BOOST model for different market segments (segmented by current
travel mode). The testing set is divided into four mutually-exclusive
subsets obtained using the current travel mode. For the overall
accuracy and for the true negative rate, i.e., the rate of
successfully predicting the ``non-switching outcome,'' the results
show that BOOST performs better for individuals who currently use Car,
Walk, and Bike; on the other hand, BOOST performs better in
successfully predicting the ``switching'' outcome (i.e., true positive
rate) for individuals who currently use Bus. Moreover, the prediction
accuracy is more balanced (i.e., true positive rate and true negative
rate are similar) for individuals using Car and Bus than individuals
using Walk and Bike. One possible explanation for the relatively lower
performance in predicting the switching behavior of those who
currently walk or bike is the low mode-switching rates of these
individuals in the training set (Walk: 20.26\%; Bike: 13.45\%), which
resulted in an imbalanced classification problem. Properly handling
the imbalanced classes in machine learning is a crucial research
topic, and is an important topic for future work.

\subsection{Partial Dependence Plots (PDPs) and Conditional Partial Dependence Plots (CPDPs)}

\begin{figure}[!t]
    \centering
    \begin{subfigure}[b]{0.48\textwidth}
        \centering
        \includegraphics[height=2.8in]{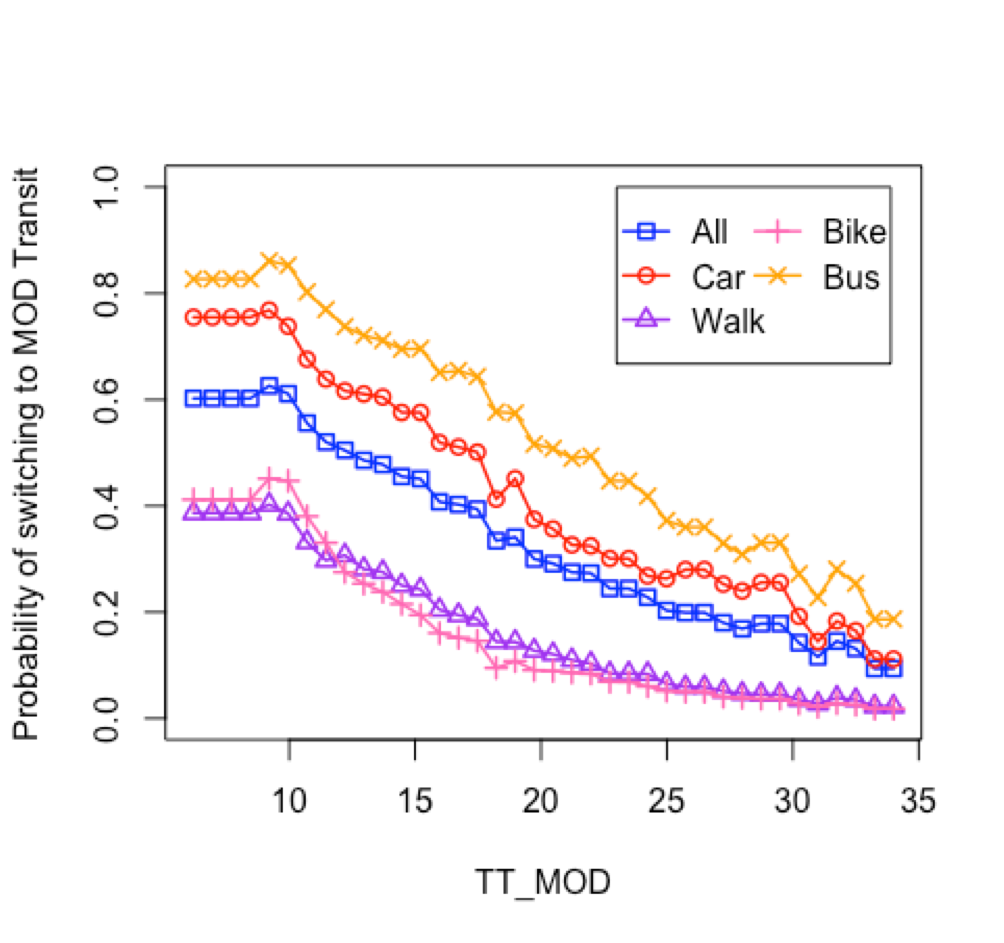}
        \caption{Partial Dependence on TT\_MOD}
        \label{fig:pdptt_pt}
    \end{subfigure}%
    ~
    \begin{subfigure}[b]{0.48\textwidth}
        \centering
        \includegraphics[height=2.8in]{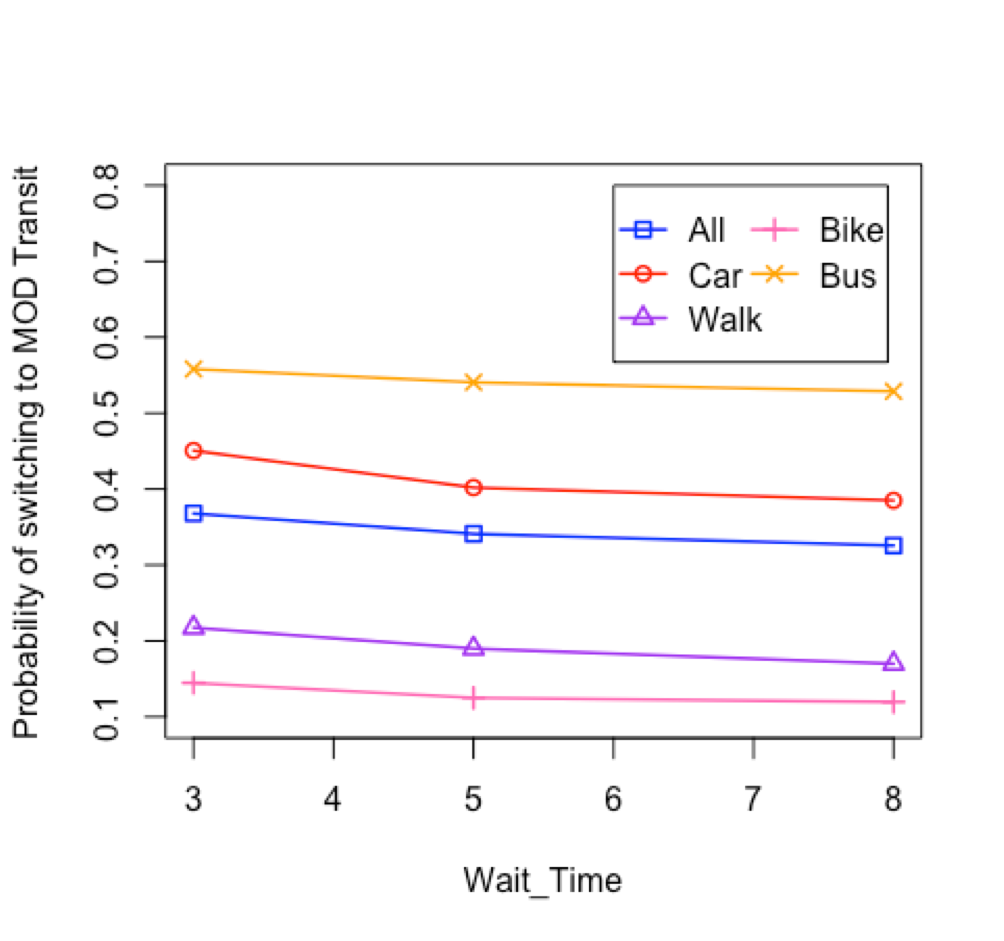}
        \caption{Partial Dependence on Wait\_Time}
        \label{fig:pdpWaittime}
    \end{subfigure}%

    \begin{subfigure}[b]{0.48\textwidth}
        \centering
        \includegraphics[height=2.8in]{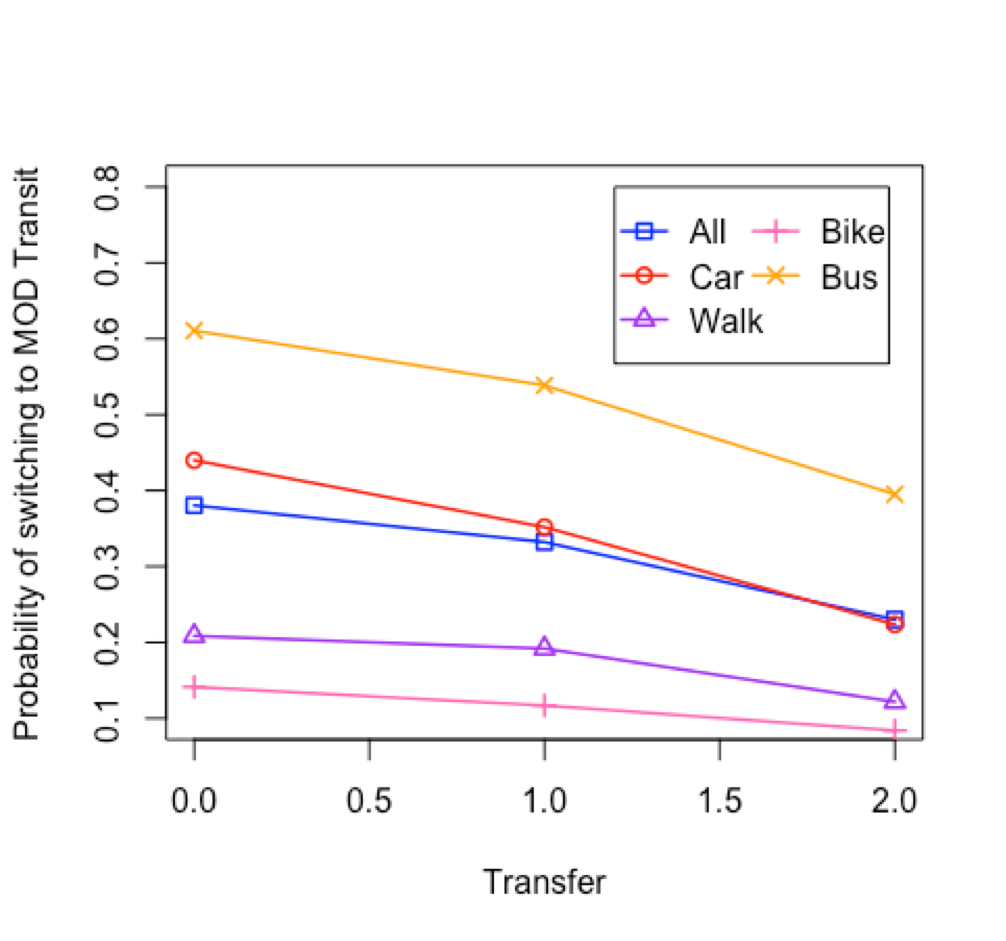}
        \caption{Partial Dependence on Transfer}
        \label{fig:pdptransfer}
    \end{subfigure}
    ~
    \begin{subfigure}[b]{0.48\textwidth}
        \centering
        \includegraphics[height=2.8in]{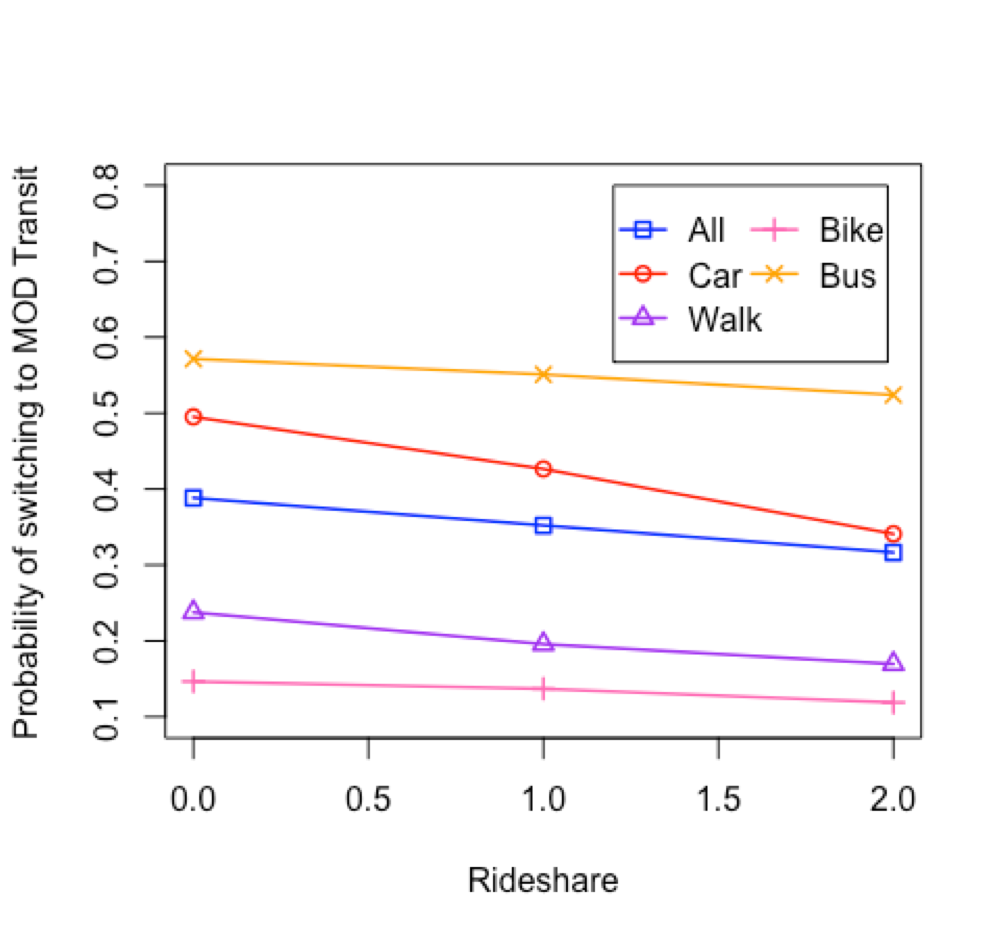}
        \caption{Partial Dependence on Rideshare}
        \label{fig:pdprideshare}
    \end{subfigure}
    \caption{PDPs and CPDPs of Level-Of-Service Variables for MOD Transit.}
    \label{fig:PDP}
\end{figure}

The PDPs and CPDPs of the level-of-service variables (i.e., TT\_MOD,
Wait\_Time, Transfer, and Rideshare) for MOD Transit are presented in
Figure \ref{fig:PDP}. The $y$-axis is the probability of switching to
MOD Transit, while the $x$-axis is a level-of-service variable for the
proposed system. The blue lines in Figure \ref{fig:PDP}
are the PDPs, while the remaining lines are CPDPs for different market
segments (Car: red; Walk: purple; Bike: pink; and Bus: yellow). 

The PDPs have the expected decreasing trend for all the
level-of-service variables: As the level-of-service for MOD Transit
gets worse, the preference for MOD Transit declines for the entire
market and for each market segment. 

For TT\_MOD (see Figure \ref{fig:pdptt_pt}), the corresponding PDPs
and CPDPs present strong nonlinearities for the entire market and the
four market segments. The probability of switching remains largely
unchanged when the transit time is less than 10 minutes. For Walk and
Bike, the switching probability decreases faster from 10 minutes to 20
minutes compared to that from 20 minutes to 34 minutes. For Car and
Bus, the switching probability present an approximately linear
decrease after 10 minutes. These results show that the MOD Transit
system is highly desirable across different market segments when the
transit time is relatively small (i.e., less than 10 minutes).  In
addition, there exists some non-smooth perturbations in the PDPs and
CPDPs, even though the general trend is decreasing. This is probably
caused by the tree structure of the BOOST model. For Wait\_Time and
Transfer, some nonlinearities are also present as illustrated in
Figure \ref{fig:pdpWaittime}-\ref{fig:pdptransfer}. To be specific,
for all the subpopulations, the switching probability has larger
decrease for Wait\_Time from 3 minutes to 5 minutes than that from 5
minutes to 8 minutes. In addition, within the analysis range for
Transfer (i.e., 0--2), the probability of switching to MOD Transit has
larger decline for the second transfer than the first transfer. One
insight is that one may consider limiting the number of transfers to
at most one when designing a new MOD Transit system.

These results demonstrate that PDPs and CPDPs are valuable tools for behavioral analysis: They
can readily reveal the nonlinear relationships between the outcome variable (i.e., switching
probabilities) and the independent variables of interest (i.e. level-of-service variables).

\begin{table}[t!]
\centering
\caption{Approximate Global Slopes for PDPs and CPDPs}
\begin{tabular}{c|cccc}
\hline
\textbf{Mode} & \textbf{TT\_MOD} & \textbf{Wait\_Time} & \textbf{Transfer} & \textbf{Rideshare} \\ \hline
\textbf{All}  & $-0.018$           & $-0.008$              & $-0.075$            & $-0.036$             \\
\textbf{Car}  & $-0.023$           & $-0.013$              & $-0.108$            & $-0.077$             \\
\textbf{Walk} & $-0.013$           & $-0.009$              & $-0.043$            & $-0.034$             \\
\textbf{Bike} & $-0.014$           & $-0.005$              & $-0.029$            & $-0.014$             \\
\textbf{Bus}  & $-0.023$           & $-0.006$              & $-0.108$            & $-0.024$             \\ \hline
\end{tabular}
\label{tab:slopes}
\end{table}

Table \ref{tab:slopes} presents the approximate global slopes of the PDPs
and CPDPs, which can be viewed as being comparable to the beta coefficient estimates
in the random utility models. The approximate slopes of Transfer and
Rideshare show that Transfer has a larger impact on the entire market
and all the population segments. Among the four market segments,
Transfer has the most impact on the existing Car and Bus users, which
indicates that vehicle riders are more sensitive and reluctant to
transfers. Moreover, the last column of Table \ref{tab:slopes} shows
that Car users are most reluctant to additional pickups. One possible
explanation is that Car users value privacy the most and are less
willing to rideshare than other population segments. Travelers that used to
ridesharing, such as Bus users, are not very sensitive to Rideshare.
The approximate slopes for TT\_MOD show that Car and Bus users have
the same preference ($-0.023$), while Walk and Bike users are similar
in their preferences ($-0.013$ and $-0.014$). The results are
intuitive since Car and Bus belong to motorized modes, while Walk and
Bike are similar in nature and often combined together into one soft
mode \citep{omrani2015predicting}.  The approximate global slopes of
TT\_MOD and Wait\_Time indicate that Wait\_Time has smaller impact on
the entire market and all the subpopulations. One possible explanation
is that on-demand shuttles in the new MOD Transit system are requested
via webpage or mobile phones: Hence Wait\_Time may be perceived as the
active waiting time, resulting in smaller penalties for MOD
Transit. In addition, existing Car users have the largest slope
($-0.013$) compared to other modes. Traveling by fixed-route bus
requires one to walk to and wait at bus stops and existing walkers and
bikers are used to traveling outside, but Car users, by contrast,
usually spend very little out-of-vehicle travel time, which may make
them more sensitive to Wait\_Time \citep{hitge2015comparison}.

\subsection{Individual Conditional Expectation (ICE) Plots and Conditional Individual Partial Dependence Plots (CIPDPs)}

\begin{figure}[!t]
    \centering
    \begin{subfigure}[b]{0.68\textwidth}
        \centering
        \includegraphics[height=3.2in]{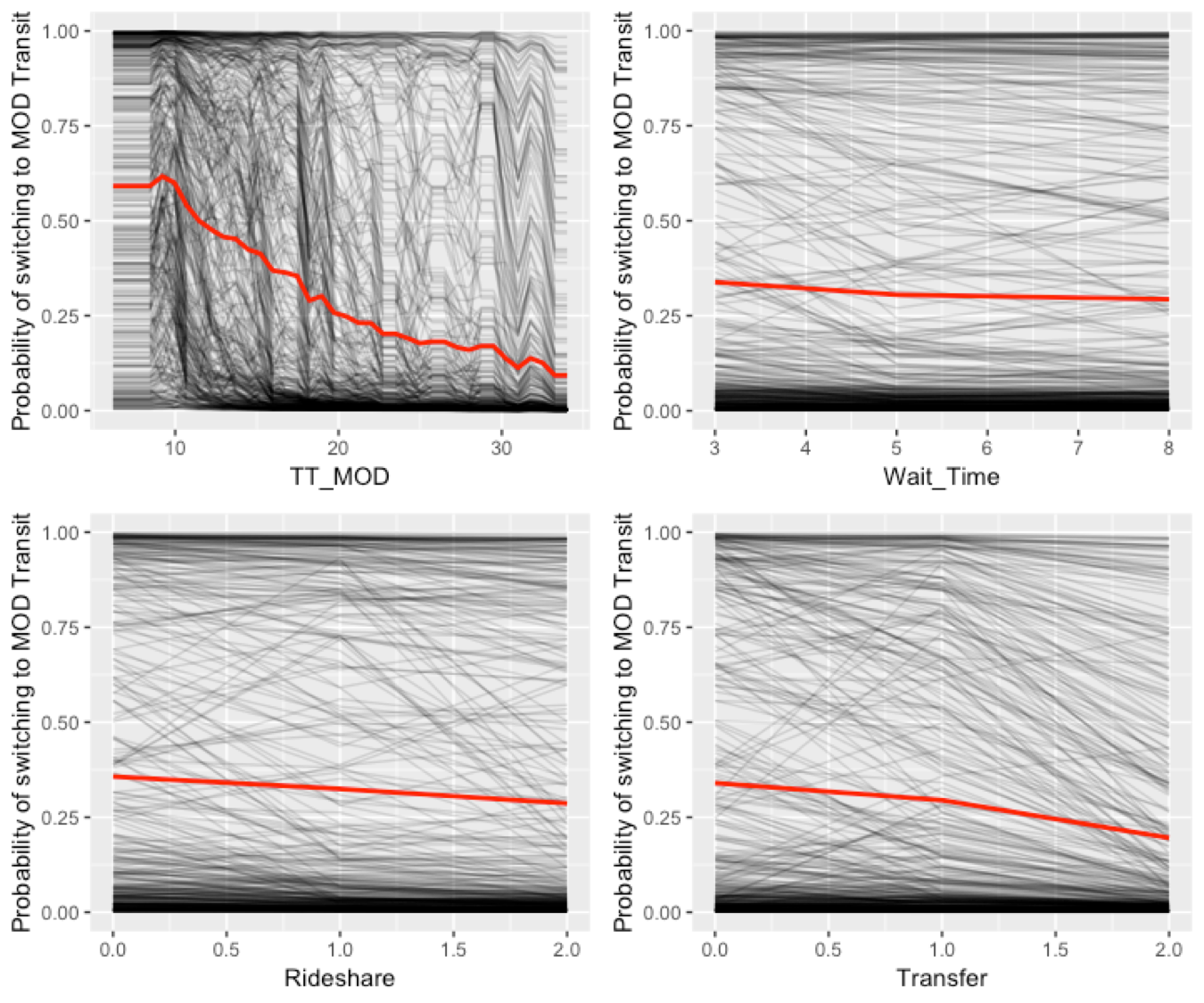}
        \caption{ICE Plots}
        \label{fig:ice_all}
    \end{subfigure}%
    
    \begin{subfigure}[b]{0.68\textwidth}
        \centering
        \includegraphics[height=3.2in]{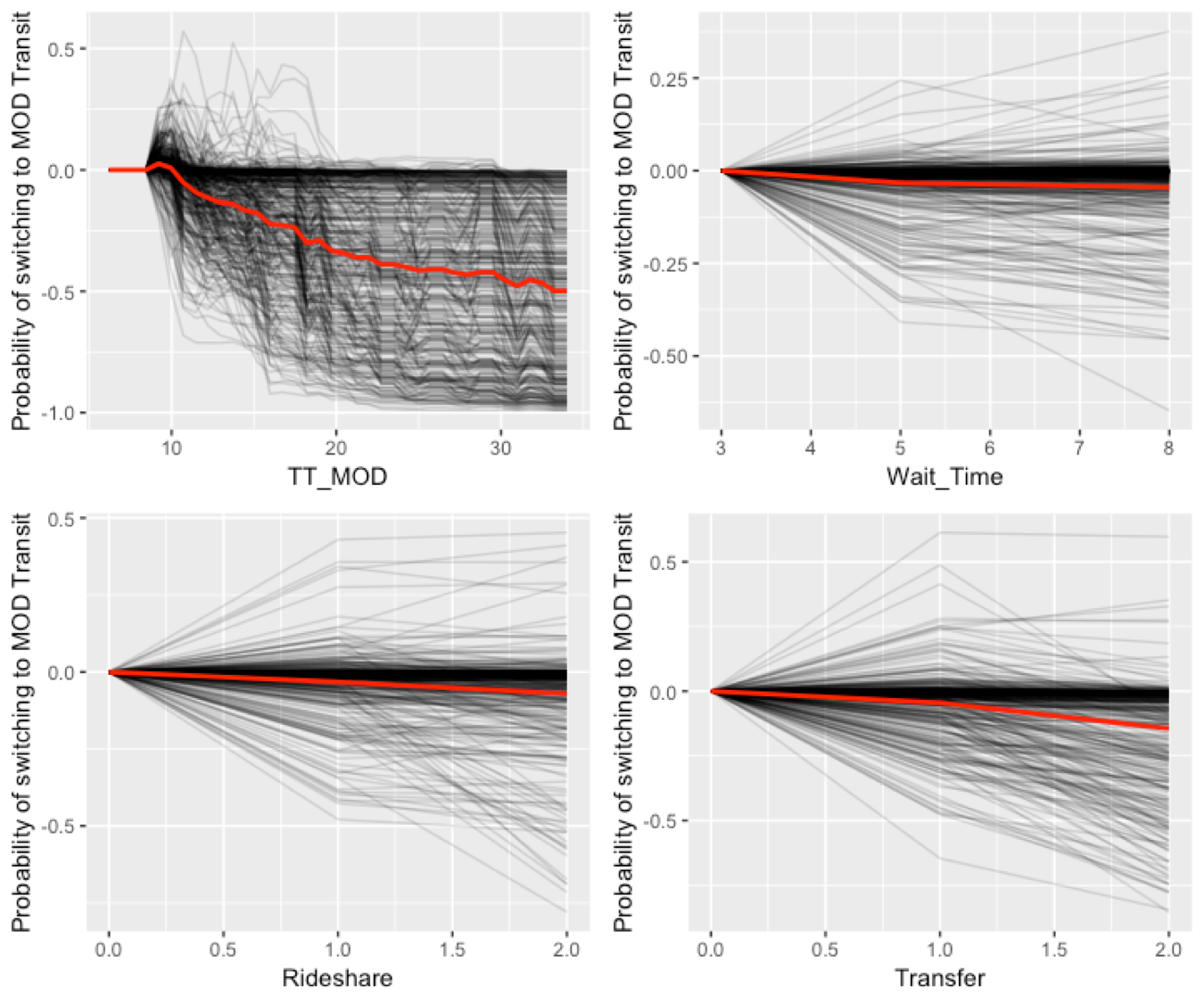}
        \caption{C-ICE Plots}
        \label{fig:ice_c}
    \end{subfigure}%
    \caption{ICE and C-ICE Plots of Level-of-Service Variables for MOD Transit: The Red Lines Represent the Corresponding PDP.}
    \label{fig:C-CIPDP}
\end{figure}

To further reveal individual taste heterogeneity (response heterogeneity to different trip attributes), we randomly sampled 100
instances from each subpopulation (by existing travel mode) and
constructs ICE plots of the level-of-service variables for MOD
Transit. The resulting 400 instances are shown in Figure
\ref{fig:ice_all}. The ICE plots show extensive response
heterogeneity. However, it is hard to conclude whether the ICE curves
differ between individuals since all the curves start at various
predictions, especially for TT\_MOD. Therefore, to uncover individual heterogeneity more reliably (and more clearly), Figure \ref{fig:ice_c} uses
centered ICE (C-ICE) plots: They center the plots around a certain
point within the range and display only the difference in the
prediction with respect to this point
\citep{goldstein2015peeking,molnar}. The resulting C-ICE plots shows
that some people are behaving very differently from the majority of
observed individuals. 

\begin{figure}[!t]
    \centering
    \includegraphics[width=16cm]{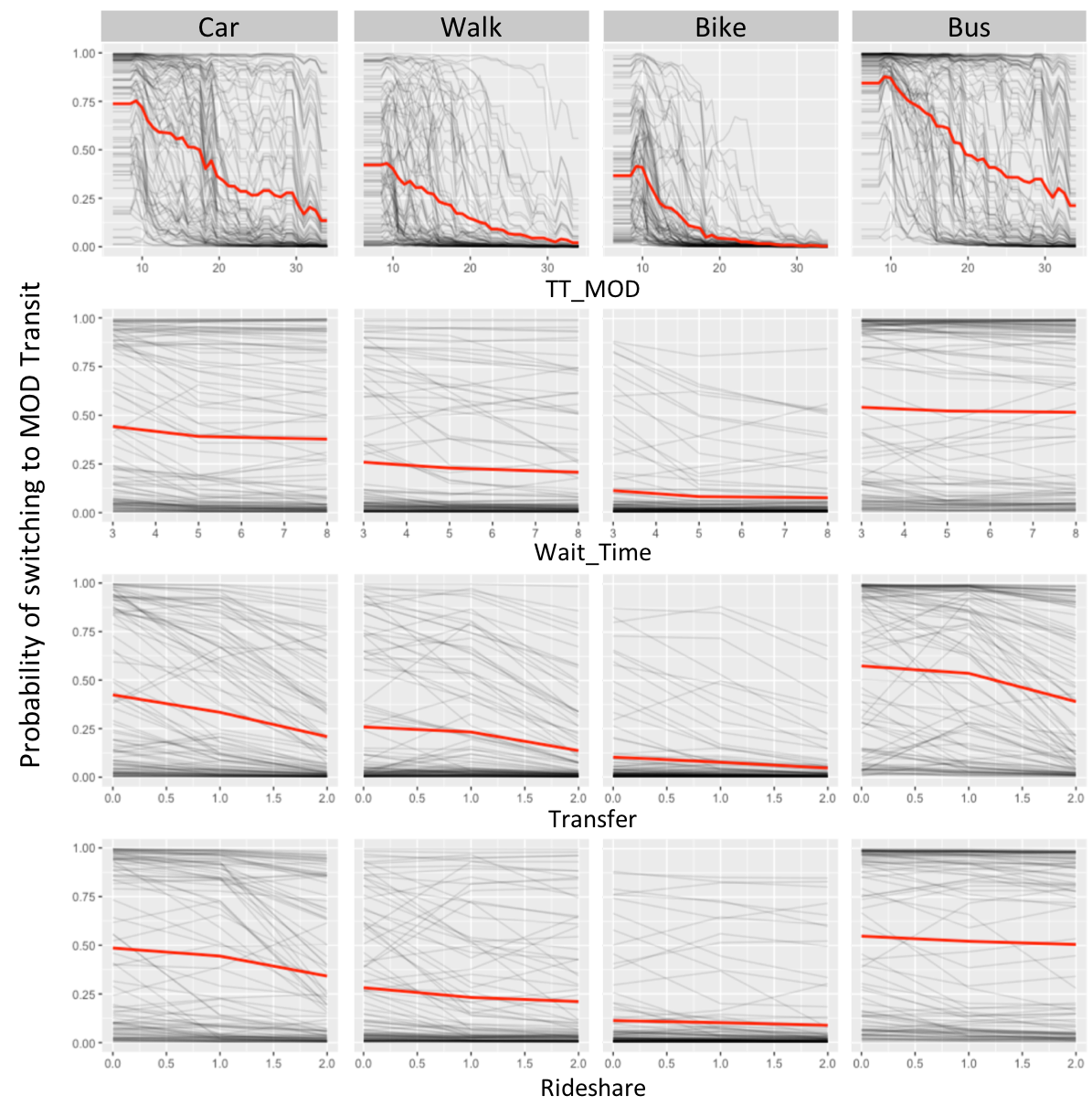}
    \caption{CIPDPs of Level-of-Service Variables for MOD Transit: The Red Lines Represent the Corresponding CPDPs.}
    \label{fig:CIPDP}
\end{figure}

\begin{figure}[!t]
    \centering
    \includegraphics[width=16cm]{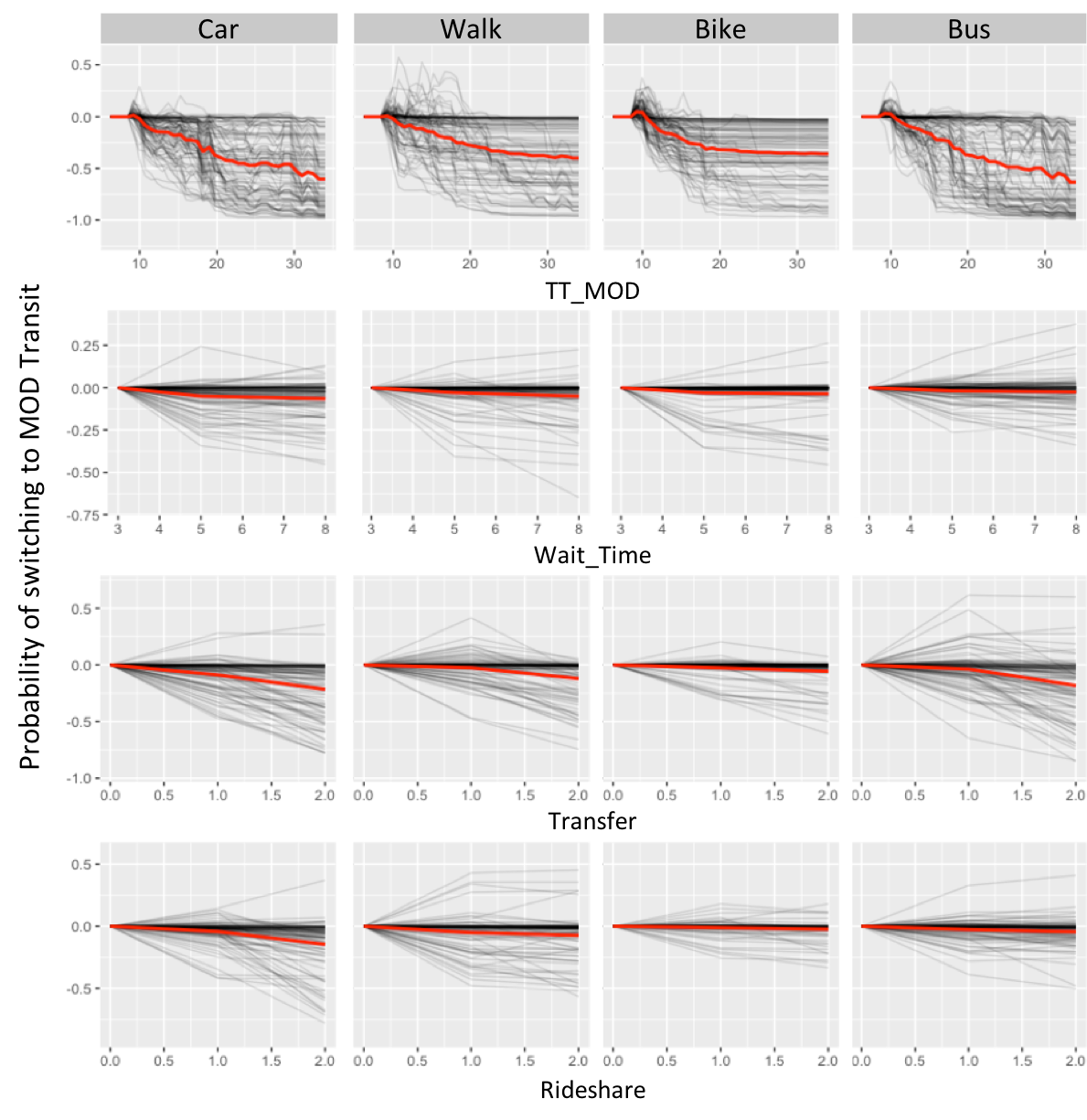}
    \caption{Centered CIPDPs of Level-of-Service Variables for MOD Transit: The Red Lines Represent the Corresponding Centered CPDPs.}
    \label{fig:CCIPDP}
\end{figure}

Figures \ref{fig:CIPDP} and \ref{fig:CCIPDP} present the CIPDPs of
different level-of-service variables for the market segments
determined by the current travel mode. For TT\_MOD, individuals in the
Car and Bus segments exhibit more homogeneous trend of switching
behavior, as all curves seem to follow the similar pattern and there are no obvious
interactions. These results show that CPDPs for these two modes
provide a good summary of the relationships between TT\_MOD and the
predicted switching probabilities well. For Walk and Bike segments,
the majority of the subpopulation is following the similar trend, but
there are a few individuals who behave differently from the rest of
the subpopulation. For example, for Walk, the switching probabilities
of several individuals do not decrease until around 30 minutes
transit time.


For Wait\_Time, all four market segments present response
heterogeneity. Many Car users have a significant decrease in switching
probability from 3 minutes to 5 minutes of waiting time, while many
others do not seem to be sensitive to waiting time at all. Many
walkers are similar in behavior to drivers with a few exceptions: They
have a consistent decrease from 3 minutes to 8 minutes Wait\_Time. For
Bike users, the results are interesting: Most travelers are lumped at
the bottom, showing strong intrinsic preference for not switching. On
the other hand, a dozen bikers have the most decrease from 3 minutes
to 5 minutes Wait\_Time. Many public transit users present a steady
decrease from 3 minutes to 8 minutes of waiting time, while a few of
them have the most decrease from 3 minutes to 5 minutes.

For Transfer and Rideshare, significant response heterogeneity exist within and
across different subpopulations. For Car users, a significant amount
of individuals have a larger decrease in switching probabilities for
the second Transfer/Rideshare, while some others have the most
decrease for the first Transfer/Rideshare. Most walkers have a
consistent decrease as Transfer changes from 0 to 2 and have the most
signifcant decrease for the first Rideshare. Most bikers present no
interest of switching to MOD Transit, with some of them showing a
consistent decrease for Transfer and Rideshare. Lastly, most Bus users
show larger decreases in switching probability on the second Transfer,
while a few of them have larger decreases for the first Transfer. In
terms of Rideshare, many display a consistent decrease, while some of
them have the most decrease for the first additional pickup.

\begin{figure}[!t]
    \centering
    \includegraphics[width=16cm]{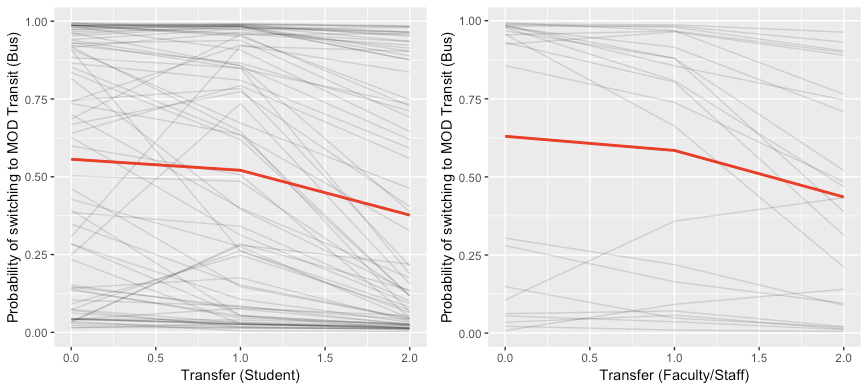}
    \caption{CIPDPs of Level-of-Service Variables for MOD Transit: Conditioning on Current Mode and Faculty/Student Status.}
    \label{fig:CIPDP-Stu-Fac}
\end{figure}


However, some individual responses seem to be unrealistic: Taking
Transfer as an example, some individuals in Walk and Bus segments
exhibit increases in switching probabilities as Transfer changes from
0 to 1, and the switching probabilities decrease significantly as
Transfer changes from 1 to 2. Figure \ref{fig:CIPDP-Stu-Fac} applies
a two-dimensional segmentation to understand this anomaly. It shows
that many more anomalies for students than faculty, indicating that
some students may have misreported some of their preferences. This also
illustrates the benefits of CPDPs and CIPDPs in zooming on the results
to build a better understanding.

With high predictive accuracy among all the market segments (over
86\%) and reasonable taste heterogeneity patterns revealed
above, it may be argued that \textit{the BOOST model segments the
  entire population and capture heterogeneity automatically}; that is to say, the BOOST model accounts for individual heterogeneity by itself without explicit model-specification efforts from the modeler. This
claim is consistent with the findings in \citet{lheritier2018airline},
which has shown that the RF model captures segments in an automated
way, by comparing RF with the latent class multinomial logit model. As
discussed in Subsection 4.2, BOOST, BAG, and RF are all tree-based
ensemble models with some differences.


\subsection{Marginal Effects and Elasticities by Current Travel Mode}

As discussed in Subsection 2.2, market segmentation is a widely-used
approach to reduce the large number of instances being dealt with into
a manageable number of mutually-exclusive groups that share
well-defined characteristics \citep{anable2005complacent}. After
applying market segmentation to identify different groups, it becomes
possible to make predictions about their responses to various
conditions and generate better-targeted policies. Therefore, this
subsection applies market segmentation to marginal effects and
elasticities (organized by the current travel mode), to evaluate how
different subpopulations react to system changes.

Table \ref{tab:ME_E} presents the marginal effects and elasticities by
current travel mode for the level-of-service variables of MOD
Transit. As discussed in Subsection 3.2.3, tree-based models (such as
the BOOST model) cannot make ``out-of-bound'' predictions: Hence only
the marginal effects and elasticities for the instances within the
data boundary are presented. For instance, when Transfer is under
evaluation, only instances with no or one transfer are considered when
computing the marginal effects for one unit increase, since no data is
available on three transfers. By contrast, when computing the marginal
effects for one unit decrease, only those instances with one or two
transfer(s) are extracted for analysis. In this analysis, Transfer,
Rideshare, and Wait\_Time\footnote{Wait\_Time only takes on three
  different values, i.e., 3, 5, and 8 minutes, in the survey.}  are
treated as discrete features and TT\_MOD is treated as a continuous
feature. Marginal effects are only computed for continuous variables,
since it is not meaningful to analyze 1\% increase or decrease for the
discrete variables. Note also that Wait\_Time uses a $-2$ unit instead
of $-1$ unit because it can only take three values separated by two
units.


\begin{table}[!t]
\caption{Marginal Effects and Elasticities for Wait\_Time, Transfer, Rideshare, and TT\_MOD.}
\centering
\begin{tabular}{cccccccc}
\hline
\multicolumn{2}{c}{Variable}       & $\delta$                                         & All     & Car     & Walk    & Bike    & Bus     \\ \hline
\multirow{2}{*}{Wait\_Time} & \multirow{2}{*}{Marginal Effects} & +1 unit & $-2.12$\% & $-2.95$\% & $-2.42$\% & $-1.79$\% & $-1.45$\% \\
                            &                                   & $-2$ unit & 0.79\%  & 1.68\%  & 0.65\%  & 0.33\%  & 0.65\%  \\ \hline
\multirow{2}{*}{Transfer}   & \multirow{2}{*}{Marginal Effects} & +1 unit & $-5.32$\% & $-8.71$\% & $-2.36$\% & $-2.72$\% & $-8.18$\% \\
                            &                                   & $-1$ unit & 8.68\%  & 9.57\%  & 5.14\%  & 2.30\%  & 10.19\% \\ \hline
\multirow{2}{*}{Rideshare}  & \multirow{2}{*}{Marginal Effects} & +1 unit & $-4.16$\% & $-8.03$\% & $-4.00$\% & $-1.90$\% & $-2.90$\% \\
                            &                                   & $-1$ unit & 3.22\%  & 7.97\%  & 2.43\%  & 0.96\%  & 2.40\%  \\ \hline
\multirow{4}{*}{TT\_MOD}    & \multirow{2}{*}{Marginal Effects} & +1 unit & $-2.60$\% & $-2.91$\% & $-2.25$\% & $-2.33$\% & $-2.94$\% \\
                            &                                   & $-1$ unit & 2.49\%  & 2.95\%  & 2.10\%  & 1.60\%  & 2.94\%  \\
                            & \multirow{2}{*}{Elasticities}       & +10\%    & $-1.44$  & $-1.62$  & $-1.93$  & $-2.88$  & $-1.07$  \\
                            &                                   & $-10$\%    & 1.54   & 1.74   & 1.85   & 2.76   & 1.25   \\ \hline
\end{tabular}
\label{tab:ME_E}
\end{table}


As shown in Table \ref{tab:ME_E}, the marginal effects of TT\_MOD are
largely consistent across different market segments. In the
literature, in-vehicle travel time was often used as a benchmark to
evaluate the penalty of other level-of-service variables
\citep{iseki2009not}. This paper however uses TT\_MOD as the benchmark
instead. The travel time of MOD Transit consists of two parts,
including in-vehicle travel time and out-of-vehicle travel time, and
the out-of-vehicle travel time is typically found to have more
negative impacts on choosing public transit, compared to the
in-vehicle travel time \citep{YAN2018}. Therefore, when using TT\_MOD
to evaluate the penalty of other variables, it is expected to have
smaller values compared to previous findings in the literature (which
often use in-vehicle time instead). In addition, compared to the
current transit users, the elasticities of TT\_MOD are larger for
individuals who are currently driving, walking, and biking to work,
showing their relatively lower preferences for switching to MOD
Transit.

The marginal effects of Wait\_Time are relatively small across
different travel modes, with Car users having the largest marginal
effects ($-2.95$\% for +1 unit and 1.68\% for $-2$ unit). As discussed
above, a possible explanation is that drivers are most sensitive to
out-of-vehicle travel time.  In addition, the results for +1 unit and
$-2$ unit display noticeable differences for marginal effects of
Wait\_Time. A possible reason is the strong nonlinearity of
Wait\_Time. In general, the penalty of Wait\_Time seems small in our
case study, which is consistent with our previous findings (see CPDP
results for Wait\_Time).


For the entire market, the marginal effect results indicate that one
transfer can be converted into 2.0--3.5 minutes MOD travel time
(including both in-vehicle and out-of-vehicle travel times). Even
though the values are relatively small, some previous studies have
similar findings (see \citet{iseki2009not}). Rideshare has relatively
smaller impacts compared to Transfer, which can be converted into
1.3--1.6 min of TT\_MOD. In other words, 1 Transfer is equivalent to 2
additional pickups, which is somewhat as expected since many people
may prefer a longer in-vehicle travel time for Rideshare to a shorter
out-of-vehicle travel time for Transfer.


On the other hand, Transfer and Rideshare present distinct impacts on
different market segments. For Car and Bus users, 1 Transfer can be
converted into 3.0--3.2 minutes of TT\_MOD (Car) and 2.8--3.5 minutes
of TT\_MOD (Bus) respectively. 1 Rideshare is equivalent to 2.7--2.8
minutes TT\_MOD for drivers and 0.8--1.0 minutes of TT\_MOD for public
transit users. The results show that existing populations who are
using motorized modes to commute are most reluctant to transfers. This
is consistent with our prior observations and may be explained by an
increased awareness about potential delays caused by transfers in
these population segments.  Moreover, Rideshare has a significant
negative impact on Car users indicating, once again, that Car users
may value privacy the most, while people used to share rides (e.g.,
Bus riders) are less affected.

By contrast, one transfer is equal to 1.0--2.4 minutes of TT\_MOD for
walkers and 1.2--1.4 minutes of TT\_MOD for bikers. One rideshare can
be converted into 1.2--1.8 minutes of TT\_MOD for walkers and 0.6--0.8
minutes TT\_MOD for bikers. These results are relatively small, and a
possible explanation is that Walk/Bike users are more likely to be
active commuters, who value sustainable transport and promote an
environmentally friendly lifestyle. As a result, transfers (involving
out-of-vehicle walking and waiting time) and rideshares (involving
carpooling) may help reduce traffic and CO$_2$ emission---aligning
with their values and promoting a more sustainable public transit
system. Therefore, walkers and bikers are not very reluctant to take
transfers or rideshares.

\subsection{Summary of Key Findings}

The application of various model-agnostic interpretation tools to a
machine-learning model with high predictive accuracy demonstrates that
significant taste heterogeneity exist within and
across different subpopulations, as illustrated by market segmentation using the current travel mode. Arguably, the machine-learning model can segment the entire
population and accommodate individual heterogeneity in an automated way.

In terms of behavioral insights, an important finding is that the existing Car and Bus users are more sensitive to transfers than pedestrians and cyclists. Another finding is that existing drivers are more reluctant to share rides with others, perhaps due to a higher preference for privacy. 

\section{Conclusion}

This paper studied the potential of machine-learning classifiers to
{\em predict} and {\em explain} travel mode switching behavior in
heterogeneous markets. It demonstrated that machine learning
significantly outperforms the logit model on a case study in terms of
prediction. Moreover, the paper demonstrated that it is reasonable to
claim that BOOST, the most accurate model in terms of out-of-sample
prediction, automatically segments the market, providing key insights
on various population segments at no cost. The paper also highlighted
that the BOOST results can be interpreted by existing model-agnostic
tools, such as PDPs and ICE plots.  Furthermore, the paper introduced
two new tools, CPDPs and CIPDPs, to extend PDPs and ICE curves to
various subgroups using market segmentation.  Results on the case
study indicated that the machine-learning classifier and the
interpretation tools may help reveal individual
heterogeneity in order to facilitate decision making.

We identify three potential directions for research research. The first
research avenue can addresses the fact that PDPs and ICE plots assume the
independence of features under evaluation from the remaining
features. However, a certain level of correlation among variables
exists in nearly every real-world dataset. Taking our dataset as an
example, TT\_MOD is somewhat correlated with other travel time
variables, since the origin-destination distance is the same. To
overcome this issue, a potential direction is to apply the accumulated
local effects (ALE) plot \citep{apley2016visualizing}, by only
visualizing how the model predictions change in a small ``window''
around a particular feature value. This idea can be applied to PDPs,
ICE plots, and their generalizations proposed in the paper. It should
lead to new model-agnostic interpretation tools to address this
independence assumption. The second direction is to use the
information gained from the PDP, ICE, CPDP, and CIPDP of black-box
models to design the utility functions of the logit models
automatically, thus enhancing their model performance. Lastly, there
was an imbalanced classification problem in our case study, and
similar problems can be found in many other previous studies that
focused on travel behavior modeling, such as
\citet{hagenauer2017comparative,xie2003work,wang2018machine}. Future
research should study and tackle this problem in a comprehensive and
systematic way.





\section*{Acknowledgements}

This research was partly funded by the Georgia Institute of
Technology, the Michigan Institute of Data Science (MIDAS), and Grant
7F-30154 from the Department of Energy.





\bibliographystyle{elsarticle-harv}
\biboptions{semicolon,round,sort,authoryear}

\bibliography{paper.bbl}

\begin{thebibliography}{43}
\expandafter\ifx\csname natexlab\endcsname\relax\def\natexlab#1{#1}\fi
\expandafter\ifx\csname url\endcsname\relax
  \def\url#1{\texttt{#1}}\fi
\expandafter\ifx\csname urlprefix\endcsname\relax\def\urlprefix{URL }\fi

\bibitem[{Anable(2005)}]{anable2005complacent}
Anable, J., 2005. `\text{Complacent Car Addicts}' or `\text{Aspiring
  Environmentalists}'? \text{Identifying} travel behaviour segments using
  attitude theory. Transport Policy 12~(1), 65--78.

\bibitem[{Apley(2016)}]{apley2016visualizing}
Apley, D.~W., 2016. Visualizing the effects of predictor variables in black box
  supervised learning models. arXiv preprint arXiv:1612.08468.

\bibitem[{Athey(2017)}]{athey2017beyond}
Athey, S., 2017. Beyond prediction: Using big data for policy problems. Science
  355~(6324), 483--485.

\bibitem[{Bhat(2000)}]{bhat2000incorporating}
Bhat, C.~R., 2000. Incorporating observed and unobserved heterogeneity in urban
  work travel mode choice modeling. Transportation Science 34~(2), 228--238.

\bibitem[{Bhat et~al.(2016)Bhat, Astroza, and Bhat}]{bhat2016allowing}
Bhat, C.~R., Astroza, S., Bhat, A.~C., 2016. On allowing a general form for
  unobserved heterogeneity in the multiple discrete--continuous probit model:
  Formulation and application to tourism travel. Transportation Research Part
  B: Methodological 86, 223--249.

\bibitem[{Breiman(1996)}]{breiman1996bagging}
Breiman, L., 1996. Bagging predictors. Machine Learning 24~(2), 123--140.

\bibitem[{Breiman(2001)}]{breiman2001random}
Breiman, L., 2001. Random forests. Machine Learning 45~(1), 5--32.

\bibitem[{Breiman(2017)}]{breiman2017classification}
Breiman, L., 2017. Classification and Regression Trees. Routledge.

\bibitem[{Cortes and Vapnik(1995)}]{cortes1995support}
Cortes, C., Vapnik, V., 1995. Support-vector networks. Machine Learning 20~(3),
  273--297.

\bibitem[{Doshi-Velez and Kim(2017)}]{doshi2017towards}
Doshi-Velez, F., Kim, B., 2017. Towards a rigorous science of interpretable
  machine learning. arXiv preprint arXiv:1702.08608.

\bibitem[{Du et~al.(2018)Du, Liu, and Hu}]{du2018techniques}
Du, M., Liu, N., Hu, X., 2018. Techniques for interpretable machine learning.
  arXiv preprint arXiv:1808.00033.

\bibitem[{Friedman(2001)}]{friedman2001greedy}
Friedman, J.~H., 2001. Greedy function approximation: A gradient boosting
  machine. Annals of Statistics, 1189--1232.

\bibitem[{Goldstein et~al.(2015)Goldstein, Kapelner, Bleich, and
  Pitkin}]{goldstein2015peeking}
Goldstein, A., Kapelner, A., Bleich, J., Pitkin, E., 2015. Peeking inside the
  black box: Visualizing statistical learning with plots of individual
  conditional expectation. Journal of Computational and Graphical Statistics
  24~(1), 44--65.

\bibitem[{Hagenauer and Helbich(2017)}]{hagenauer2017comparative}
Hagenauer, J., Helbich, M., 2017. A comparative study of machine learning
  classifiers for modeling travel mode choice. Expert Systems with Applications
  78, 273--282.

\bibitem[{Hastie et~al.(2001)Hastie, Tibshirani, and
  Friedman}]{friedman2001elements}
Hastie, T., Tibshirani, R., Friedman, J., 2001. The Elements of Statistical
  Learning. Vol.~1. Springer series in statistics New York, NY, USA:.

\bibitem[{Hitge and Vanderschuren(2015)}]{hitge2015comparison}
Hitge, G., Vanderschuren, M., 2015. Comparison of travel time between private
  car and public transport in cape town. Journal of the South African
  Institution of Civil Engineering 57~(3), 35--43.

\bibitem[{Ho(1998)}]{ho1998random}
Ho, T.~K., 1998. The random subspace method for constructing decision forests.
  IEEE Transactions on Pattern Analysis and Machine Intelligence 20~(8),
  832--844.

\bibitem[{Iseki and Taylor(2009)}]{iseki2009not}
Iseki, H., Taylor, B.~D., 2009. Not all transfers are created equal: Towards a
  framework relating transfer connectivity to travel behaviour. Transport
  Reviews 29~(6), 777--800.

\bibitem[{Lh{\'e}ritier et~al.(2018)Lh{\'e}ritier, Bocamazo, Delahaye, and
  Acuna-Agost}]{lheritier2018airline}
Lh{\'e}ritier, A., Bocamazo, M., Delahaye, T., Acuna-Agost, R., 2018. Airline
  itinerary choice modeling using machine learning. Journal of Choice
  Modelling.

\bibitem[{Li et~al.(2016)Li, Miwa, Morikawa, and Liu}]{li2016incorporating}
Li, D., Miwa, T., Morikawa, T., Liu, P., 2016. Incorporating observed and
  unobserved heterogeneity in route choice analysis with sampled choice sets.
  Transportation Research Part C: Emerging Technologies 67, 31--46.

\bibitem[{Liaw and Wiener(2002)}]{RF}
Liaw, A., Wiener, M., 2002. Classification and regression by
  \text{randomForest}. R News 2~(3), 18--22.
\newline\urlprefix\url{http://CRAN.R-project.org/doc/Rnews/}

\bibitem[{Liu et~al.(2011)Liu, Zhang, and Wu}]{liu2011hard}
Liu, Y., Zhang, H.~H., Wu, Y., 2011. Hard or soft classification? large-margin
  unified machines. Journal of the American Statistical Association 106~(493),
  166--177.

\bibitem[{Mah{\'e}o et~al.(2017)Mah{\'e}o, Kilby, and Van~Hentenryck}]{TS2017}
Mah{\'e}o, A., Kilby, P., Van~Hentenryck, P., 2017. Benders decomposition for
  the design of a hub and shuttle public transit system. Transportation
  Science.

\bibitem[{McCallum et~al.(1998)McCallum, Nigam,
  et~al.}]{mccallum1998comparison}
McCallum, A., Nigam, K., et~al., 1998. A comparison of event models for naive
  bayes text classification. In: AAAI-98 Workshop on Learning for Text
  Categorization. Vol. 752. Citeseer, pp. 41--48.

\bibitem[{Meyer et~al.(2017)Meyer, Dimitriadou, Hornik, Weingessel, and
  Leisch}]{e1071}
Meyer, D., Dimitriadou, E., Hornik, K., Weingessel, A., Leisch, F., 2017.
  e1071: Misc Functions of the Department of Statistics, Probability Theory
  Group (Formerly: E1071), TU Wien. \text{R} package version 1.6-8.
\newline\urlprefix\url{https://CRAN.R-project.org/package=e1071}

\bibitem[{Molnar(2018)}]{molnar}
Molnar, C., 2018. Interpretable Machine Learning.
  https://christophm.github.io/interpretable-ml-book/,
  \url{https://christophm.github.io/interpretable-ml-book/}.

\bibitem[{Murdoch et~al.(2019)Murdoch, Singh, Kumbier, Abbasi-Asl, and
  Yu}]{murdoch2019interpretable}
Murdoch, W.~J., Singh, C., Kumbier, K., Abbasi-Asl, R., Yu, B., 2019.
  Interpretable machine learning: definitions, methods, and applications. arXiv
  preprint arXiv:1901.04592.

\bibitem[{Omrani(2015)}]{omrani2015predicting}
Omrani, H., 2015. Predicting travel mode of individuals by machine learning.
  Transportation Research Procedia 10, 840--849.

\bibitem[{Quinlan(2014)}]{quinlan2014c4}
Quinlan, J.~R., 2014. C4. 5: Programs for Machine Learning. Elsevier.

\bibitem[{{R Core Team}(2018)}]{R}
{R Core Team}, 2018. R: A Language and Environment for Statistical Computing. R
  Foundation for Statistical Computing, Vienna, Austria.
\newline\urlprefix\url{https://www.R-project.org/}

\bibitem[{Ridgeway(2017)}]{gbm}
Ridgeway, G., 2017. gbm: Generalized Boosted Regression Models. \text{R}
  package version 2.1.3.
\newline\urlprefix\url{https://CRAN.R-project.org/package=gbm}

\bibitem[{Ripley(2016)}]{tree}
Ripley, B., 2016. tree: Classification and Regression Trees. \text{R} package
  version 1.0-37.
\newline\urlprefix\url{https://CRAN.R-project.org/package=tree}

\bibitem[{Srinivasan and Mahmassani(2003)}]{srinivasan2003analyzing}
Srinivasan, K.~K., Mahmassani, H.~S., 2003. Analyzing heterogeneity and
  unobserved structural effects in route-switching behavior under atis: a
  dynamic kernel logit formulation. Transportation Research Part B:
  Methodological 37~(9), 793--814.

\bibitem[{Tang et~al.(2015)Tang, Xiong, and Zhang}]{tang2015decision}
Tang, L., Xiong, C., Zhang, L., 2015. Decision tree method for modeling travel
  mode switching in a dynamic behavioral process. Transportation Planning and
  Technology 38~(8), 833--850.

\bibitem[{Venables and Ripley(2002)}]{stats}
Venables, W.~N., Ripley, B.~D., 2002. Modern Applied Statistics with S, 4th
  Edition. Springer, New York, iSBN 0-387-95457-0.
\newline\urlprefix\url{http://www.stats.ox.ac.uk/pub/MASS4}

\bibitem[{Vij and Walker(2016)}]{vij2016and}
Vij, A., Walker, J.~L., 2016. How, when and why integrated choice and latent
  variable models are latently useful. Transportation Research Part B:
  Methodological 90, 192--217.

\bibitem[{Wager and Athey(2018)}]{wager2018}
Wager, S., Athey, S., 2018. Estimation and inference of heterogeneous treatment
  effects using random forests. Journal of the American Statistical Association
  113~(523), 1228--1242.
\newline\urlprefix\url{https://doi.org/10.1080/01621459.2017.1319839}

\bibitem[{Wahba(2002)}]{wahba2002soft}
Wahba, G., 2002. Soft and hard classification by reproducing kernel hilbert
  space methods. Proceedings of the National Academy of Sciences 99~(26),
  16524--16530.

\bibitem[{Wang and Ross(2018)}]{wang2018machine}
Wang, F., Ross, C.~L., 2018. Machine learning travel mode choices: Comparing
  the performance of an extreme gradient boosting model with a multinomial
  logit model. Transportation Research Record: Journal of the Transportation
  Research Board.

\bibitem[{Xie et~al.(2003)Xie, Lu, and Parkany}]{xie2003work}
Xie, C., Lu, J., Parkany, E., 2003. Work travel mode choice modeling with data
  mining: decision trees and neural networks. Transportation Research Record:
  Journal of the Transportation Research Board~(1854), 50--61.

\bibitem[{Yan et~al.(2018)Yan, Levine, and Zhao}]{YAN2018}
Yan, X., Levine, J., Zhao, X., 2018. Integrating ridesourcing services with
  public transit: An evaluation of traveler responses combining revealed and
  stated preference data. Transportation Research Part C: Emerging
  Technologies.
\newline\urlprefix\url{http://www.sciencedirect.com/science/article/pii/S0968090X18310398}

\bibitem[{Zhao and Hastie(2017)}]{zhao2017causal}
Zhao, Q., Hastie, T., 2017. Causal interpretations of black-box models. Journal
  of Business \& Economic Statistics, to appear.

\bibitem[{Zhao et~al.(2018)Zhao, Yan, Yu, and
  Van~Hentenryck}]{zhao2018modeling}
Zhao, X., Yan, X., Yu, A., Van~Hentenryck, P., 2018. Modeling stated preference
  for mobility-on-demand transit: A comparison of machine learning and logit
  models. arXiv preprint arXiv:1811.01315.

\end{thebibliography}







\end{document}